\documentclass{article}
\usepackage{arxiv}
\usepackage{lineno,hyperref}
\modulolinenumbers[5]
\usepackage{microtype}
\usepackage{subfigure}
\usepackage{booktabs} % for professional tables
\usepackage[colorinlistoftodos]{todonotes}
\usepackage{amsfonts}       % blackboard math symbols
\usepackage{graphicx} 
\usepackage{amsmath}
\usepackage{xcolor}
\usepackage{pifont}
\usepackage[utf8]{inputenc}

\newtheorem{theorem}{theorem}

\newcommand{\vtau}{{\boldsymbol{\tau}}}
\newcommand{\vx}{{\mathbf{x}}}
\newcommand{\vy}{{\mathbf{y}}}
\newcommand{\vk}{{\mathbf{k}}}

\newcommand{\vs}{{\mathbf{s}}}
\newcommand{\bbR}{{\mathbb{R}}}

\newcommand{\gp}{{\mathcal{GP}}}
\newcommand{\compO}{{\mathcal{O}}}
\newcommand{\N}{{\mathcal{N}}}
\newcommand{\F}{{\mathcal{F}}}

\newcommand{\vgamma}{{\boldsymbol{\gamma}}}
\newcommand{\freq}[1]{\hat{#1}} % frequency domain
\newcommand{\tb}[1]{\textbf{#1}} % frequency domain

\newcommand{\Finv}{{{\F}_{s\rightarrow \tau}^{-1}}}

\newcommand{\var}{{{\sigma}^2}}
\newcommand{\varn}{{\sigma}^2_{n}}
\newcommand{\Var}{{\Sigma}}

\newcommand{\tra}{^{\top}}
\newcommand{\vmu}{{\boldsymbol{\mu}}}

\newcommand{\imagi}{\jmath}
\newcommand{\uptext}{\overbrace}

\newcommand{\lsm}{\text{SLSM}}
\newcommand{\sm}{\text{SM}}
\newcommand{\lkp}{\text{LKP}}

\title{Gaussian Processes with Skewed Laplace Spectral Mixture Kernels for Long-term Forecasting}

\author{
    Kai Chen \\
              Institute for Computing and Information Sciences, Radboud University, Nijmegen, The Netherland \\
            Future Network of Intelligence Institute (FNii), The Chinese University of Hong Kong, Shenzhen, China\\
            \texttt{kyoungchen@gmail.com}
    \AND
    Twan van Laarhoven \\
              Institute for Computing and Information Sciences, Radboud University, Nijmegen, The Netherland \\
\texttt{tvanlaarhoven@cs.ru.nl}	
    \AND
    Elena Marchiori  \\
              Institute for Computing and Information Sciences, Radboud University, Nijmegen, The Netherland \\
\texttt{elenam@cs.ru.nl}
}

\begin{document}
\maketitle

\begin{abstract}
Long-term forecasting involves predicting a horizon that is far ahead of the last observation. It is a problem of high practical relevance, for instance for companies in order to decide upon expensive long-term investments. Despite the recent progress and success of Gaussian Processes (GPs) based on Spectral Mixture kernels, long-term forecasting remains a challenging problem for these kernels because they decay exponentially at large horizons. This is mainly due to their use of a mixture of Gaussians to model spectral densities. Characteristics of the signal important for long-term forecasting can be unravelled by investigating the distribution of the Fourier coefficients of (the training part of) the signal, which is non-smooth, heavy-tailed, sparse, and skewed. The heavy tail and skewness characteristics of such distributions in the spectral domain allow to capture long-range covariance of the signal in the time domain. Motivated by these observations, we propose to model spectral densities using a Skewed Laplace Spectral Mixture (SLSM) due to the skewness of its peaks, sparsity, non-smoothness, and heavy tail characteristics. By applying the inverse Fourier Transform to this spectral density we obtain a new GP kernel for long-term forecasting. In addition, we adapt the lottery ticket method, originally developed to prune weights of a neural network, to GPs in order to automatically select the number of kernel components. Results of extensive experiments, including a multivariate time series, show the beneficial effect of the proposed SLSM kernel for long-term extrapolation and robustness to the choice of the number of mixture components.  
\end{abstract}

% ============================================================================================
\setcounter{section}{0}
\section{Introduction}\label{sec:intro}
% ============================================================================================

Gaussian Processes (GPs) are an elegant Bayesian approach to modeling an unknown function and representing its predictive uncertainty. They provide regression models where a posterior distribution over the unknown function is maintained as evidence is accumulated. This allows GPs to learn involved functions when a large amount of evidence is available, and makes them robust against overfitting in the presence of little evidence \cite{Rasmussen2010,Rasmussen2006}. A GP can model a large class of phenomena through the choice of its kernel, which characterizes one's assumption about the autocovariance of the unknown function.
The choice of the kernel is a core step when designing a GP, since the posterior distribution can significantly vary for different kernels.  

\begin{figure}[thb]
    \centering
    \includegraphics[width=70mm]{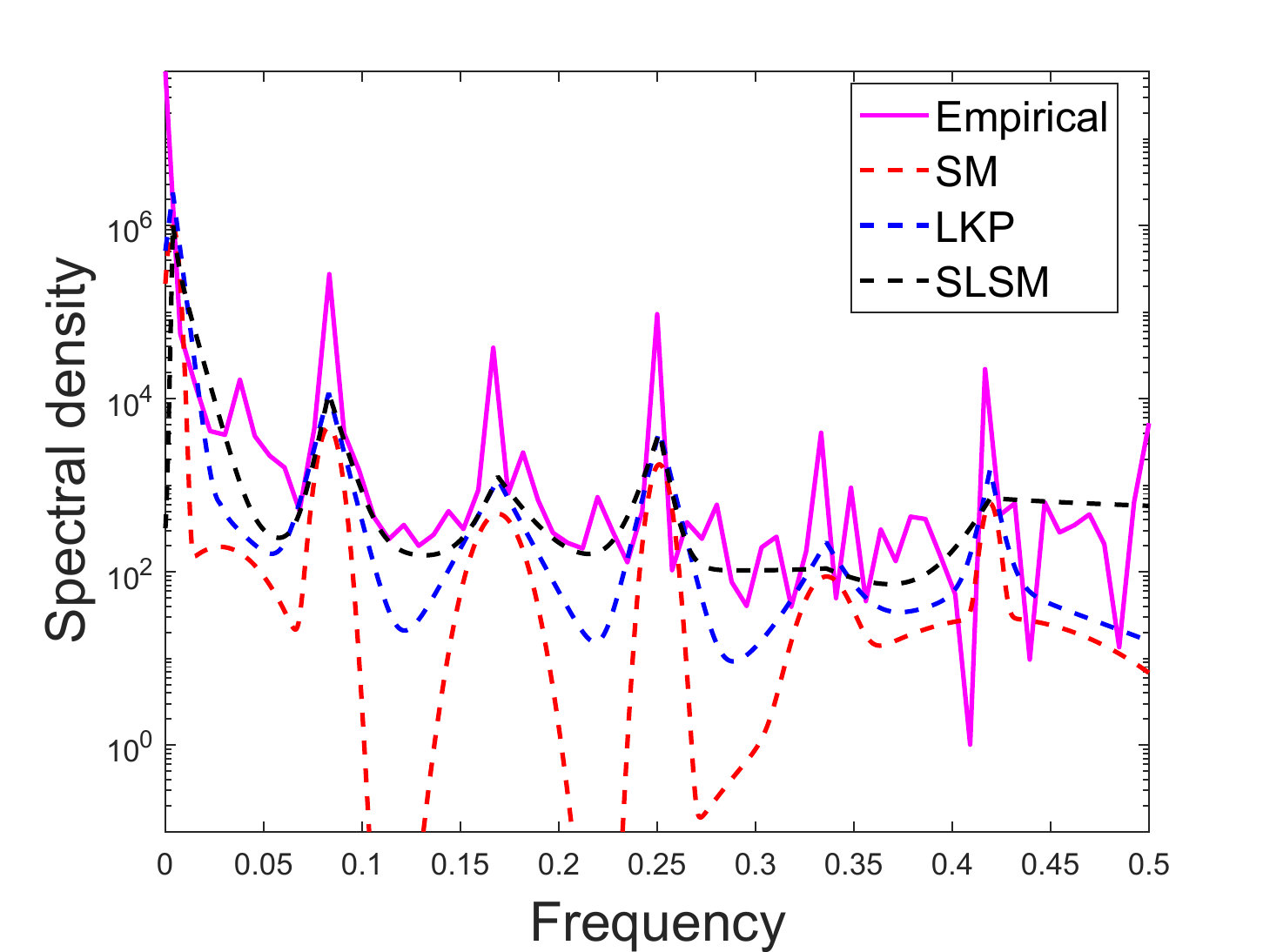}
    \label{fig:lsm-init-spec}
    \caption{Empirical spectral density (magenta solid line) of the rail passenger miles (training) time series dataset described in Section \ref{sec:exp_rail}.
        SM kernel (red dashed line), LKP kernel (blue dashed), and SLSM kernel (black dashed) fit (obtained with the least squares method) of this empirical spectral density.
    }
\end{figure}

Important advances in the design of kernel functions for GPs have been achieved in recent years. In particular, kernels based on exponential distributions  \cite{Wilson2013,Wilson2014,Duvenaud2013,Remes2017,Jang2017},  notably the flexible Spectral Mixture (SM) kernels, have been successfully applied to many real-life extrapolation tasks. However, it remains challenging for SM kernels to perform long-term forecasting. Long-term forecasting involves predictive horizons that are far ahead of the observed training data.

Our experimental analysis indicates that characteristics of the signal important for long-term forecasting can be unravelled by analyzing the distribution of the Fourier coefficients of the signal, as shown in Figure~\ref{fig:lsm-init-spec} for the (training part of the) Rail Passenger Miles time series dataset (fully described in Section \ref{sec:exp_rail}). The distribution of the Fourier coefficients  is non-smooth, heavy-tailed, sparse, and contains skewed peaks.  
\begin{figure*}
    \centering
        \subfigure{\includegraphics[width=39mm]{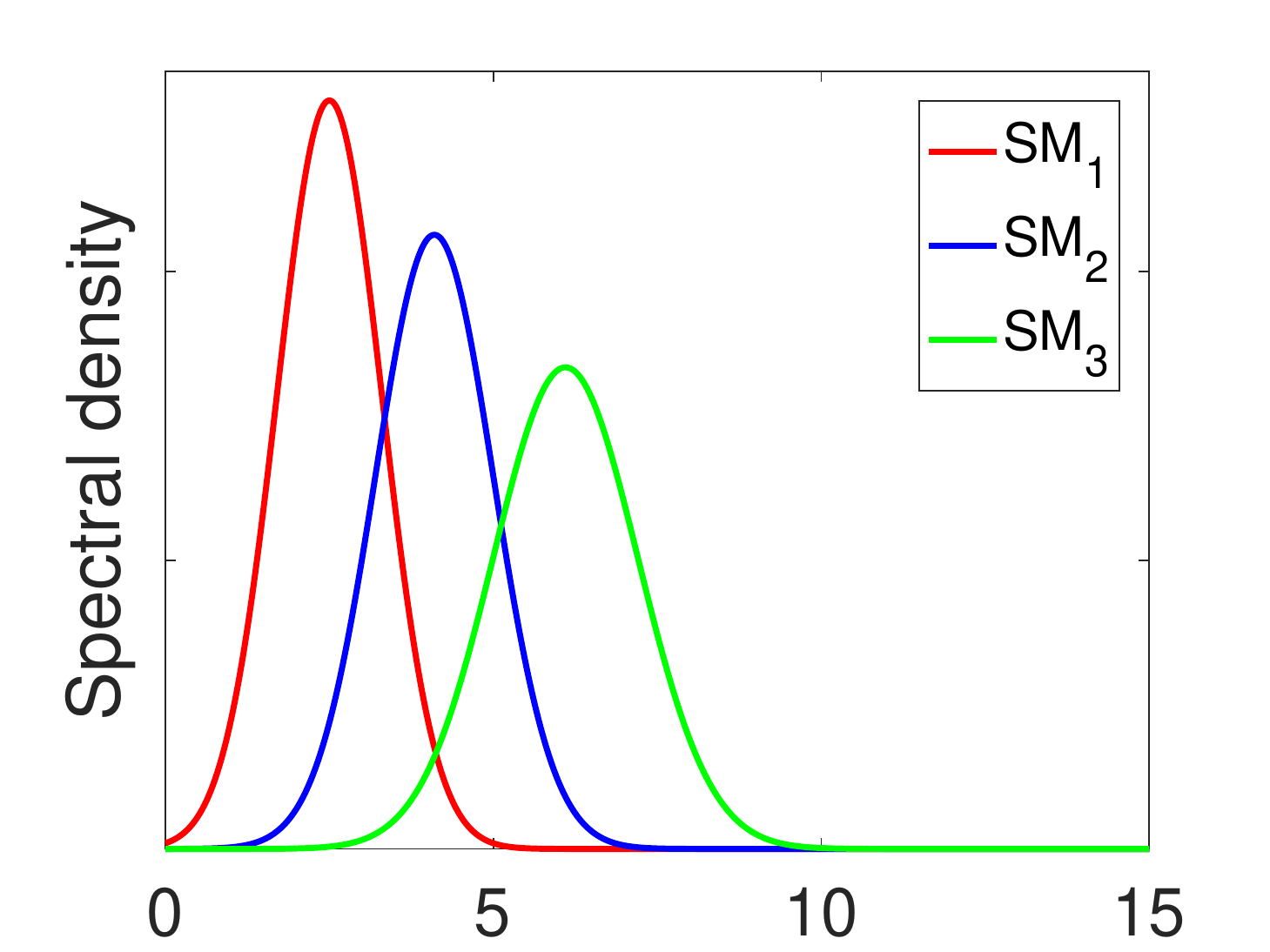}}
        \subfigure{\includegraphics[width=39mm]{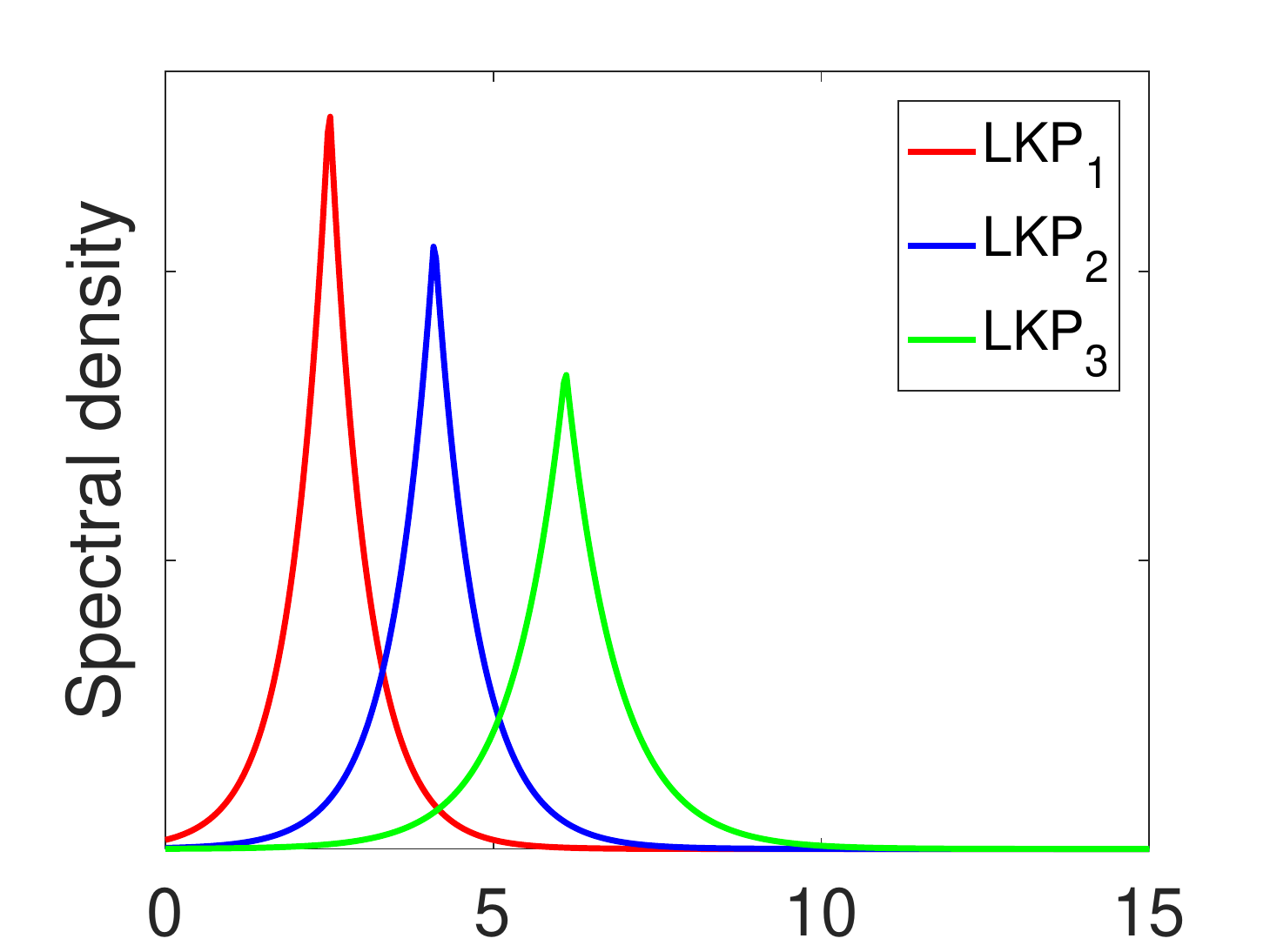}}
        \subfigure{\includegraphics[width=39mm]{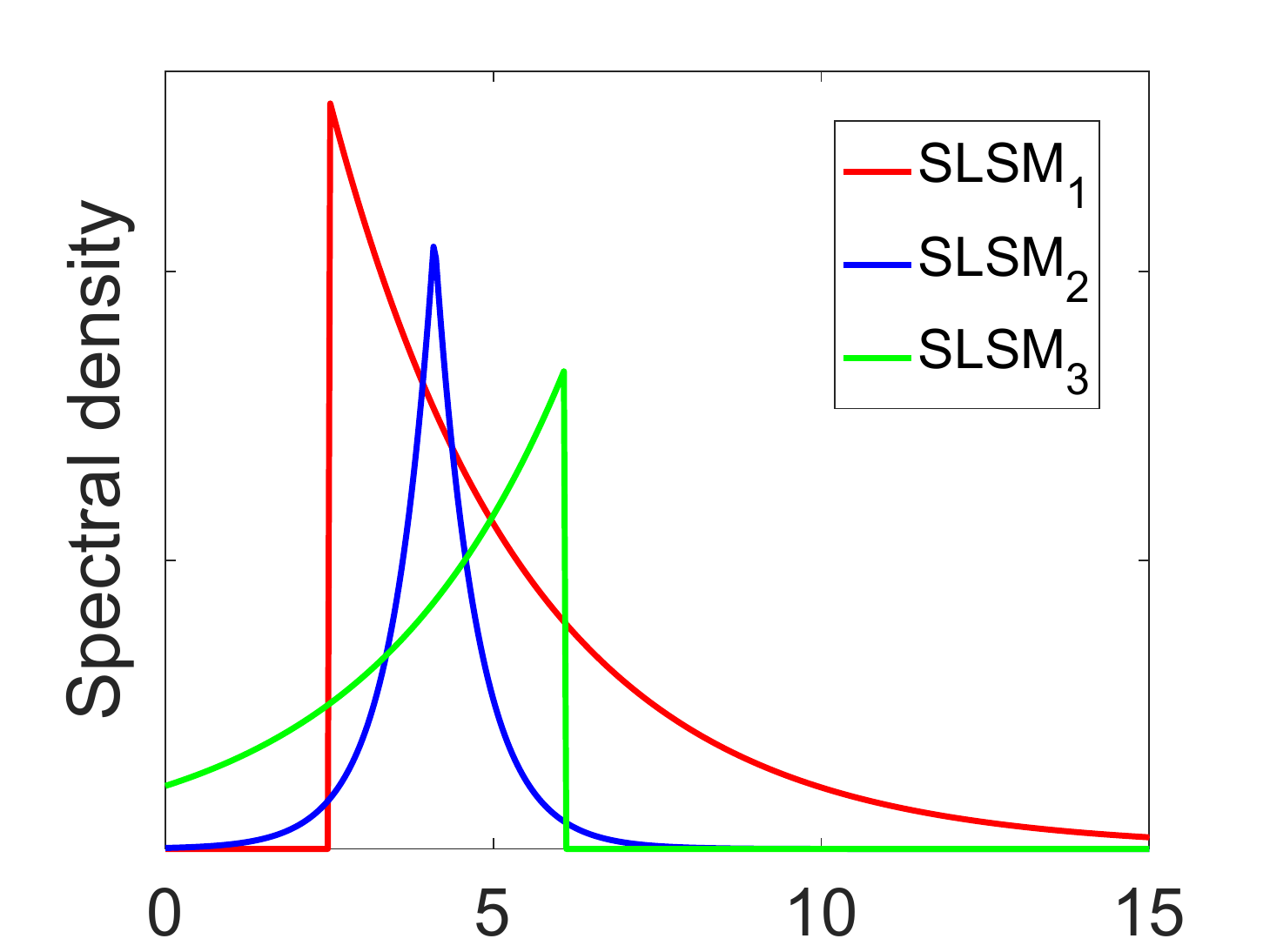}}\\
        \subfigure{\includegraphics[width=39mm]{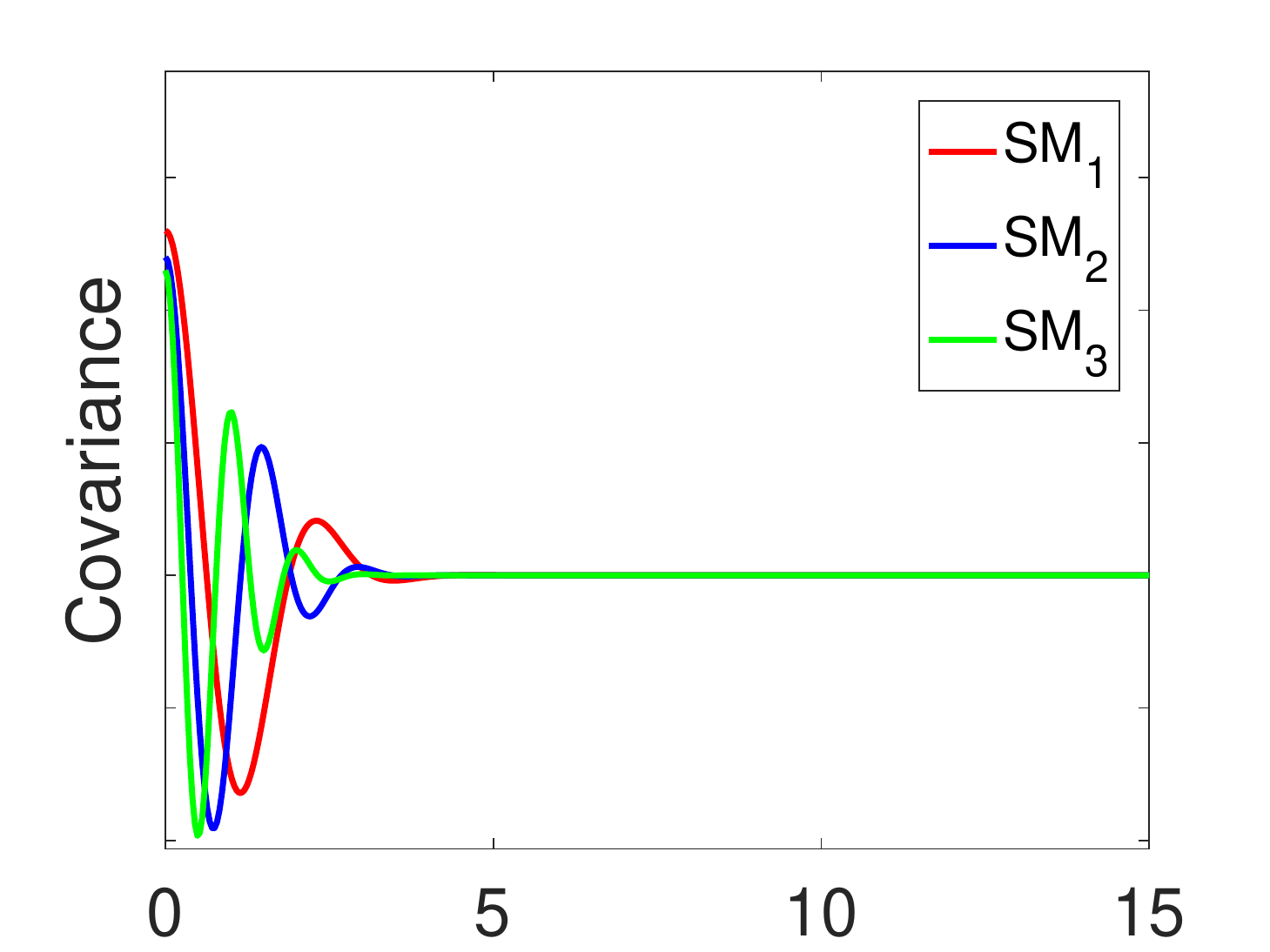}}
        \subfigure{\includegraphics[width=39mm]{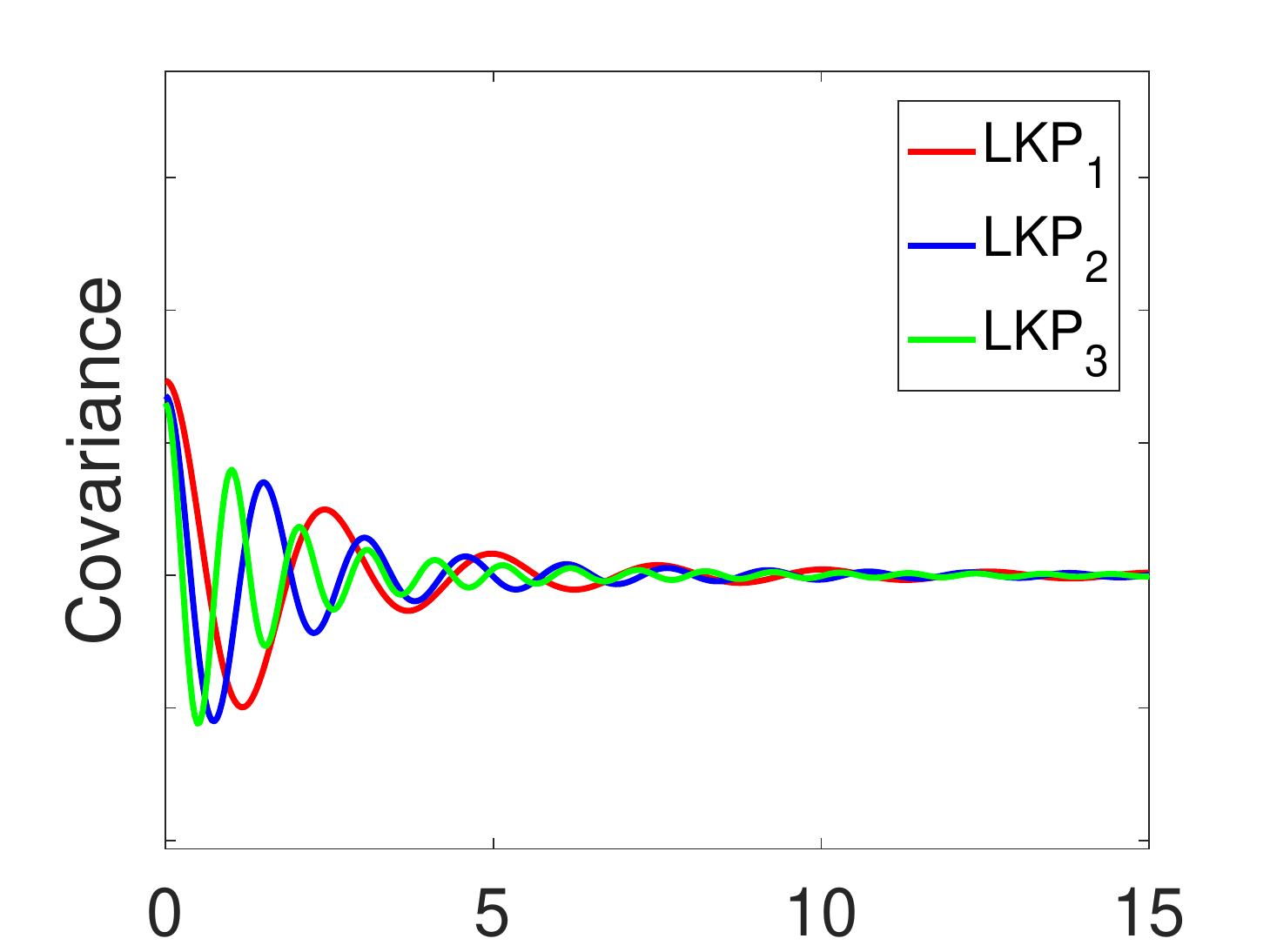}}
        \subfigure{\includegraphics[width=39mm]{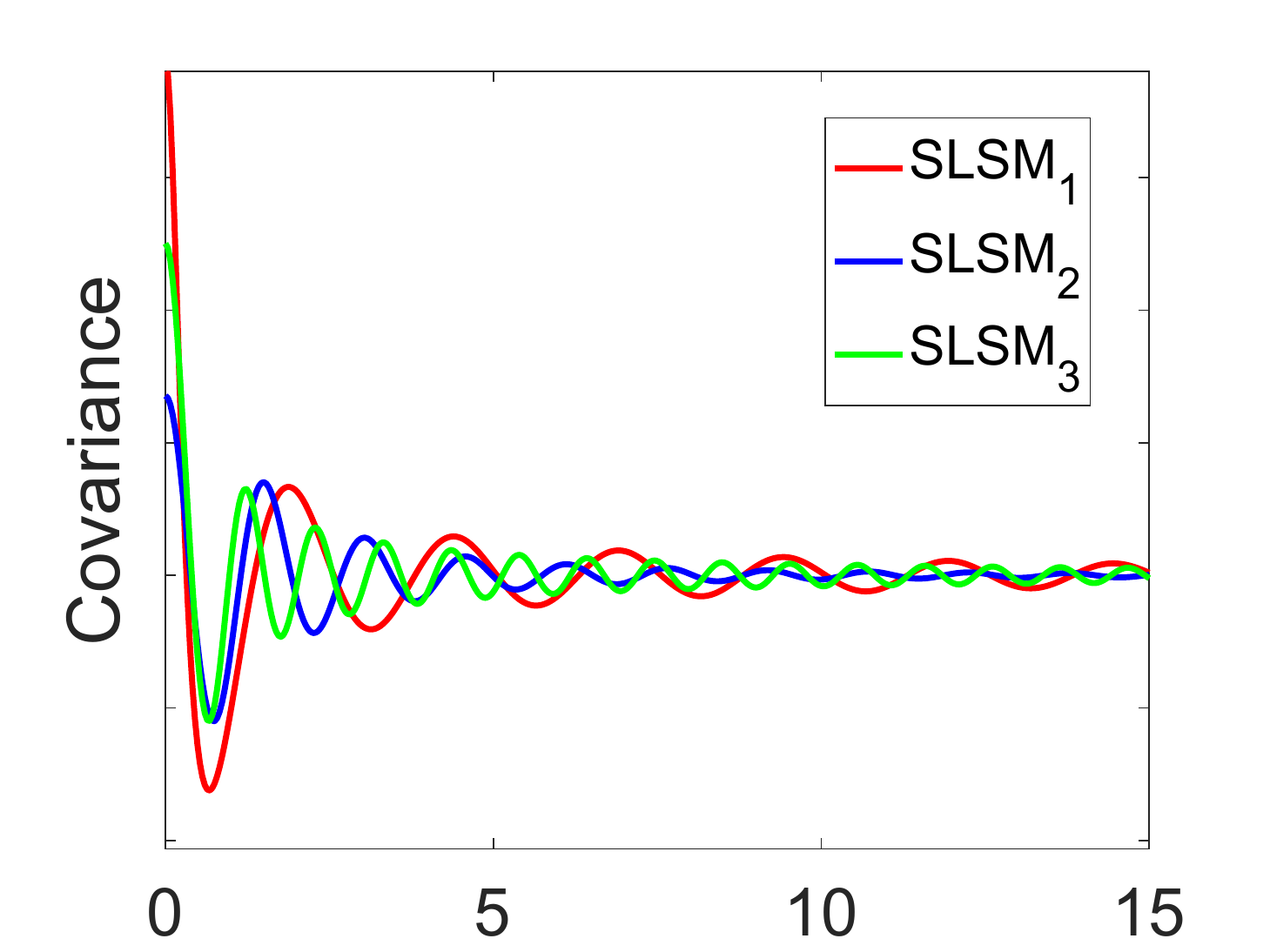}}\\
        \subfigure{\includegraphics[width=39mm]{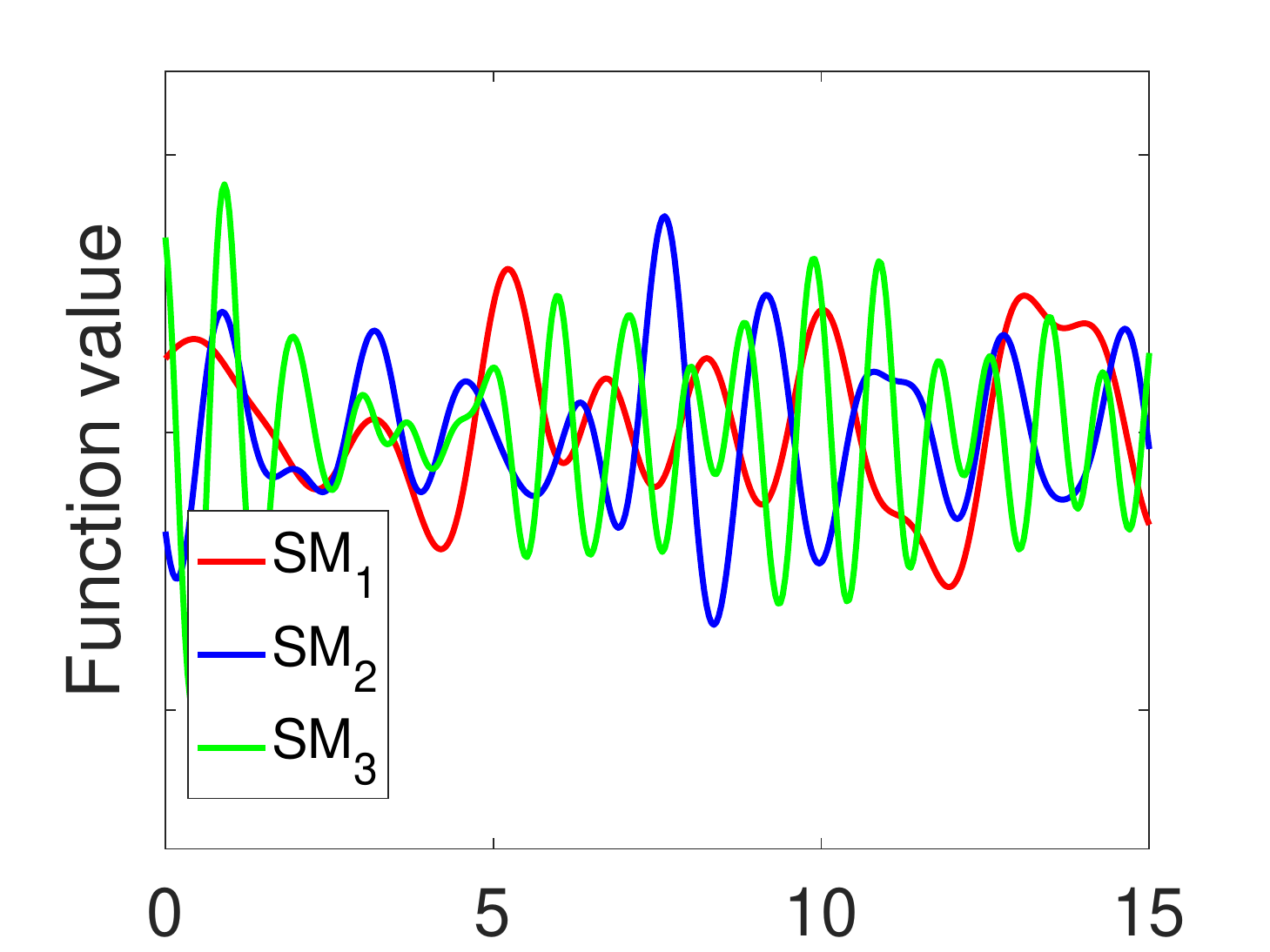}}
        \subfigure{\includegraphics[width=39mm]{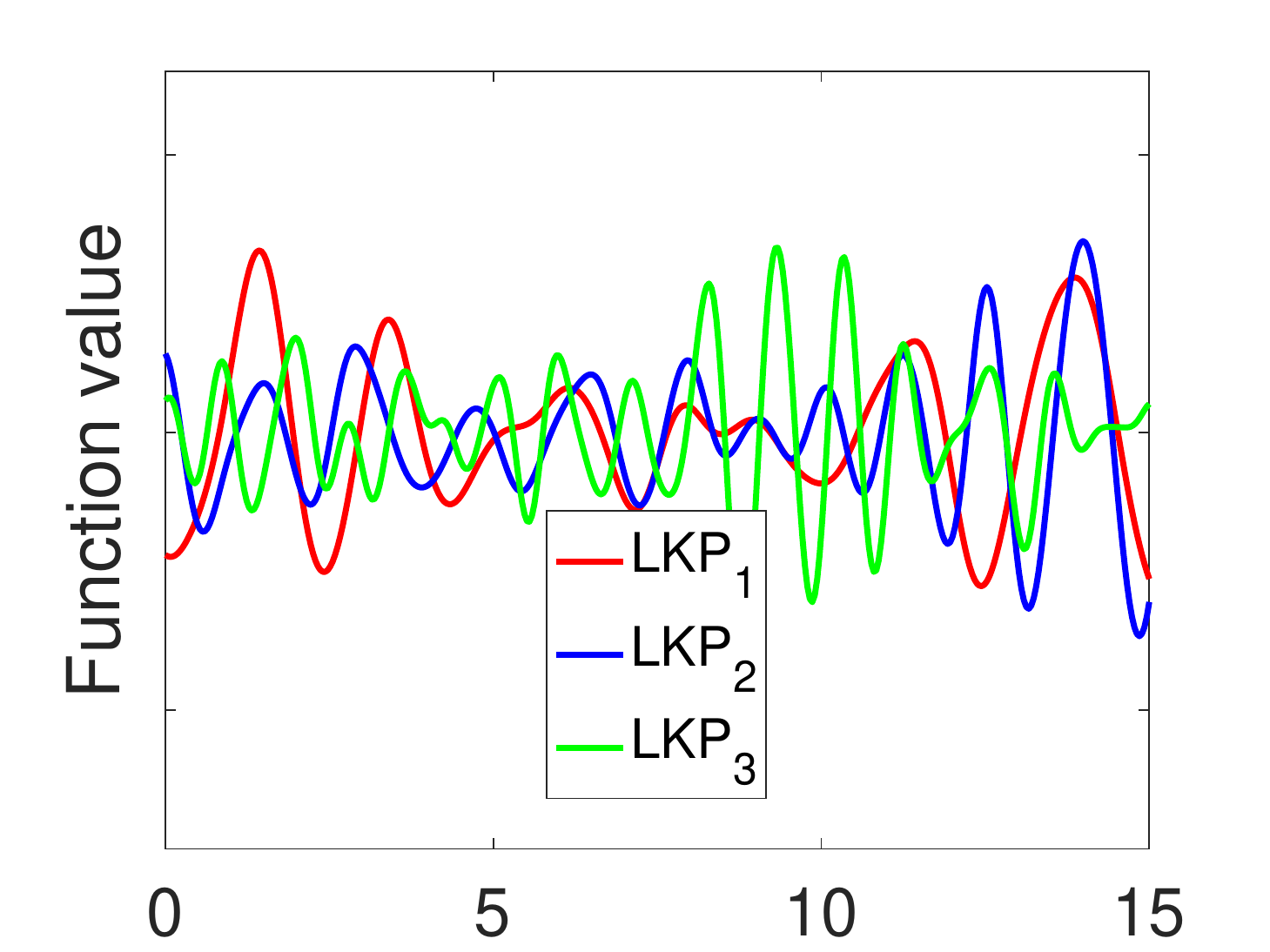}}
        \subfigure{\includegraphics[width=39mm]{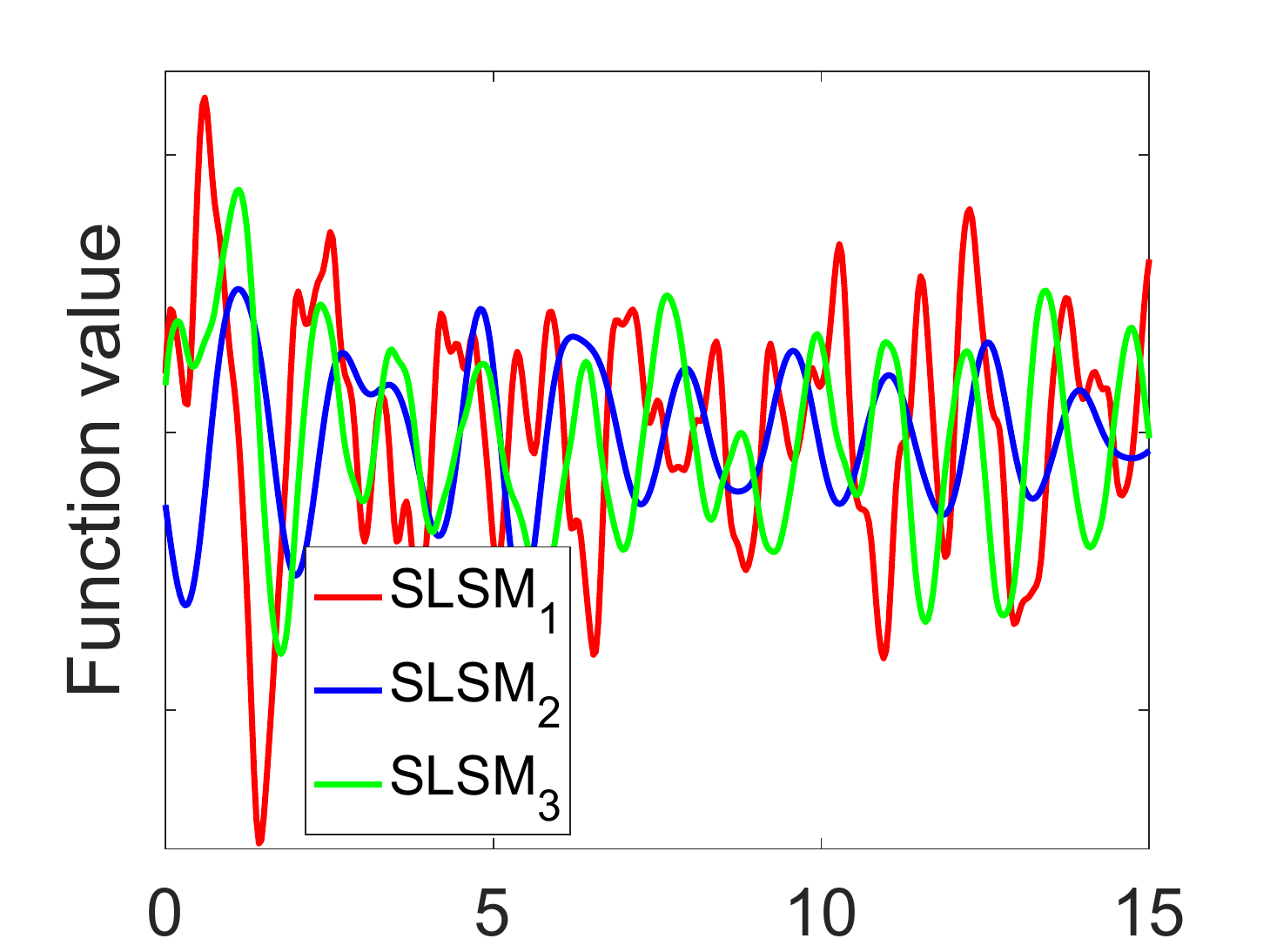}}
    \caption{
        SM (left), LKP (middle), and SLSM (right) kernels with three different choices of amplitude, mean and variance.
        Top row: different shapes of the spectral density.
        Middle row: covariance functions.
        Bottom row: 
        samples drawn from  Gaussian Processes with these kernels.        
        }\label{fig:lsm-spec}
\end{figure*}
Skewness of the peaks in the empirical density can be observed: the first, third, and fourth peak are right tailed. The heavy tails and skewness of peaks in the empirical spectral density correspond to long-range covariance in the time domain. 
We have fitted an SM kernel (red dashed line), L\'evy kernel process (LKP) kernel (blue dashed), and SLSM kernel (black dashed) on this empirical spectral density, using the least squares method.  SM and LKP cannot capture one-side heavy tail characteristics of the spectral density peaks, while SLSM can, yielding a better fit. Note that differences between the kernels are large, although in the figure they look negligible because an exponential scale is used.

Let us analyze the differences between these three spectral kernels in more detail. The inverse Fourier transform of an SM spectral density is an exponential function, so in the time domain, we have an exponential decay of covariance. When we use LKP  to fit the empirical spectral density and apply the inverse Fourier transform,  we get a Cauchy function, which has a much slower decay than that of the exponential function.
Using SLSM we extend the tails of the Laplace distribution (by means of the $\gamma$ term in Equation \eqref{eq:laplace}), which in the time domain yields a further reduction of the decay rate of the covariance. These observations are illustrated in  Figure~\ref{fig:lsm-spec}. 

In the top row, we can see the shapes of the spectral density for the three kernels, and the flexibility provided by the skewness parameter $\gamma$ in the SLSM kernel. 
Covariance functions are shown in the middle row. They indicate that SM has a very short range covariance (range $\tau<5$). LKP has longer range covariance (range $\tau< 15$) because of the heavier tails in the spectral density. SLSM has the longest covariance representation range ($\tau >>15$) due to skewness in the frequency domain.

In the bottom row, samples drawn from Gaussian Processes with these kernels are plotted. The SLSM kernel (red line) results in a larger variance, which is more flexible and less smooth than SM and LKP, because of heavy tails at high frequency position in the frequency domain. 
Both positive and negative skewness (determined by the value of $\gamma$) in SLSM can extend the covariance range and therefore increase the variance in the sampled functions.
A positive value of $\gamma$ contributes more to the signal variance.
The above observations justify the usefulness of an asymmetric Laplace kernel, capable to capture characteristics of the signal important for long-range forecasting.

Building on previous work on modeling the asymmetric tails of discrete Fourier transform coefficients distributions \cite{Gazor2003,Eltoft2006,Minguillon2001}, we investigate the use of the skewed Laplace mixture to model the spectral density, that is, the Fourier transform (FT) of a kernel. We then use the inverse Fourier transform to construct a kernel for GPs, which we call the Skewed Laplace Spectral Mixture (SLSM) kernel. Results of experiments indicate improved performance of GPs with the SLSM kernel and robustness to overfitting (when considering many mixture components). 

Our contributions can be summarized as follows:
\begin{itemize}
    \item we show how long-term forecasting with GPs is linked to long-range characteristics of covariance between random vectors, as determined by properties of the spectral density, namely 
    sparsity, skewed peaks, long tails, and non-smoothness; 
    \item we propose a new GP kernel for long-term forecasting, which includes a term  to model skewness of each spectral peak;
    \item we compare SLSM with other spectral kernels, and show that SLSM is linked to the popular Rational Quadratic kernel;  
    \item we introduce an algorithm for pruning kernel components and show experimentally its beneficial effect;
    \item we perform extensive long-term forecasting experiments with various time series data, including multivariate and large-scale ones.
\end{itemize}

The rest of the paper is organized as follows.
In the next section we summarize related works. Section~\ref{sec:gp} and~\ref{sec:sm} introduce GPs and the SM kernel.
Then in Section~\ref{sec:laplace} we introduce the SLSM kernel. Hyper-parameters initialization and automatic pruning of components are introduced in Section~\ref{sec:auto-init}. Section~\ref{sec:exp} describes experiments on real-world datasets. Concluding remarks and future work are given in Section~\ref{sec:conclude}.

% ============================================================================================
\section{Related work}\label{sec:rw}
% ============================================================================================
Because of the fundamental role of kernels in GPs, various works tackled the problem of kernel design, using two main approaches: the compositional approach, where kernels are constructed compositionally from a small set of simple base kernels, and the spectral representation approach, where kernels are constructed by modeling a spectral density, the Fourier transform of a kernel, with a Gaussian mixture, with theoretical support from the Bochner's theorem. 
In \cite{pmlr-v28-duvenaud13} the kernel learning problem is formalised as structure discovery.  A space of kernel structures is defined compositionally in terms of sums and products of a small number of base kernel structures. This provides an expressive modeling language that concisely captures many widely used techniques for constructing kernels. An algorithm that searches over this space using a greedy search is proposed. Recent works based on this approach include \cite{sun2018differentiable}, which introduces a flexible family of kernels represented by a neural network, whose architecture is based on the composition
rules for kernels, so that each unit of the network
corresponds to a valid kernel; and \cite{pearce2020expressive}, which derives Bayesian Neural Network architectures mirroring  compositional kernel combinations. The SLSM kernel with pruning strategy as proposed in this paper can be interpreted as a form of structure discovery, where unimportant components representing irrelevant structure are removed while important components representing the true underlying structure are kept during learning.

Spectral mixture kernels were introduced in \cite{Wilson2013} and are described in detail in the next section. 
The  success of these kernels has stimulated research on their extensions and improvements.  For instance, the spectral mixture product kernel \cite{Wilson2014a} was introduced  for 
modeling Cartesian structured datasets. Various works address the scalability of GPs with spectral kernels. In particular,
in \cite{Oliva2016} a Bayesian non-parametric kernel-learning is proposed, which places a non-parametric prior on the spectral distribution of random frequencies allowing it to both learn kernels and scale to large datasets. 
The limitations of stationary  kernels, notably their generalization to the non-stationary setting has been addressed in various works. For instance, the non-stationary SM kernel allows to model input-dependent GPs \cite{Remes2017}.
In \cite{shen2019harmonizable} a family of non-stationary kernels, called harmonizable mixture kernels, is introduced, which supports stationary and many non-stationary covariances. A robust kernel learning framework for this kernel family is proposed. Spectral kernels with desirable properties, called generalized spectral kernels,  have been proposed in \cite{samo2015generalized}:  they can approximate any (possibly non-stationary) bounded kernel, and  allow inference of the degree of differentiability of the corresponding stochastic process when used as a covariance function.  The SLSM kernel is stationary, a limitation discussed in the conclusion section. In \cite{chen2019incorporating}  a new kernel called generalized convolution spectral mixture is proposed, which uses cross  convolution to model dependencies between components of a spectral kernel. The SLSM kernel does not explicitly model dependencies between components.

Our investigation differs from previous works mainly because of the focus on long-term forecasting. The above-mentioned spectral kernels consider a short tailed and smooth Gaussian as a base representation of the spectral density, which corresponds to an exponential function for the covariance in the time domain. Therefore these kernels are less suited for long-term forecasting.

In this respect, the most related work is 
the L\'evy kernel process (LKP) \cite{Jang2017}. LKP was used for modeling the sharply peaked spectral density with a location-scale mixture of Laplace components, which is more sparse than the SM kernel, hence more robust to the choice of the number of mixture components. LKP can discover heavy-tailed patterns which are useful for long-range extrapolation. However it is only available for one dimensional inputs and does not model skewness characteristics of the spectral density investigated in the proposed SLSM, which, as demonstrated by our extensive experiments, is beneficial for long-term forecasting.
Specifically, SLSM  extends the tails of the Laplace distribution (by means of the  $\gamma$ term in Equation \eqref{eq:laplace}), which in the time domain yields a slower decrease of the decay rate of the covariance.  The value of $\gamma$ gives increasing weight to the right  ($\lim{\gamma \to 1}$) or to the left ($\lim{\gamma \to -1}$) tail of the density.

In general, the Laplace distribution has been successfully used in the past within diverse signal processing applications. Here we mention few examples. In \cite{bhowmick2006laplace} a Laplace mixture model is proposed, as a long-tailed alternative to the Normal distribution when identifying differentially expressed genes in micro-array experiments, and provide an extension to asymmetric over- or under-expression by means of an asymmetric extension of the mixture model.  
In \cite{Eltoft2006}  the multivariate Laplace probability model in the context of a Normal variance mixture model is analyzed and an application of  the model to represent the statistics of the discrete Fourier transform coefficients of a speech signal is given. 
In \cite{Gazor2003} samples of a speech signal are considered as
samples of a random variable, and the speech signal's probability density function in the time domain is investigated. It is shown that the distribution of speech
samples is well described by a Laplace distribution. In \cite{HyunJinPark2004} a method for capturing nonlinear dependencies in images of natural scenes is introduced, which builds a hierarchical model based on Independent Component Analysis and a mixture of Laplace distributions.
These works mainly focus on the representation and modeling of spectral density using the Laplace distribution. Instead, our focus is on kernel design for GPs.

% ============================================================================================
\section{Background}
\subsection{Gaussian processes}\label{sec:gp}
% ============================================================================================

A Gaussian process defines a distribution over functions, specified by its mean function $m({\vx})$ and covariance function $k(\vx, \vx')$ for given input vectors ${\vx},\vx'\in{\bbR}^{P}$ \cite{Rasmussen2006}. Thus we can define a GP as
\begin{align}
f({\vx})\sim{\gp}(m({\vx}), k({\vx}, {{\vx}{'}})).
\end{align}
Without loss of generality we assume the mean of a GP to be zero. The covariance function is used to construct a positive definite covariance matrix on the set $X$ of training points, called the kernel and denoted by $K=k(X, X)$.

By placing a GP prior over functions through the choice of a kernel and parameter initialization, from the training data $X$ we can predict the unknown function value $\tilde{y}_*$ and its variance $\mathbb{V}[\tilde{y}_*]$ (that is, its uncertainty) for a test point ${\vx}_*$ using the following key predictive equations for GP regression \cite{Rasmussen2006}:
\begin{subequations}
\begin{align}
    \tilde{y}_*&=\vk_{*}\tra(K+{{\varn}}I)^{-1}{\vy}\\
    \mathbb{V}[{\tilde{y}_*}]&=k({\vx}_*, {\vx}_*)-\vk_{*}\tra(K+{{\varn}}I)^{-1}\vk_{*},
    \label{eq:gp_pred}
\end{align}
\end{subequations}
where $\vk_{*} = k(\vx_*,X)$ is the covariance vector between ${\vx}_*$ and $X$, and ${\vy}$ are the observed function values corresponding to $X$.
Typically, kernels contain free hyper-parameters, denoted by $\Theta$, which can be optimized by minimizing the Negative Log Marginal Likelihood (NLML) of the observed values:
\begin{equation}\label{eq:nlml}
\begin{split}
\text{NLML}&=-\log\ p({\vy}|{X},{\Theta})\\
&\propto\uptext{\frac{1}{2}{\vy}\tra(K+{{\varn}}I)^{-1}{\vy}}^{\text{model fit}}+ \uptext{\frac{1}{2}\log|K + {{\varn}} I|}^{\text{complexity penalty}}
\end{split}
\end{equation}
where ${{\varn}}$ is the noise level and NLML is obtained through marginalization over the latent function \cite{Rasmussen2006}. This formulation follows directly from the fact that ${\vy}\sim {\N}(0, K+{{\varn}}I)$.

In this paper we consider only stationary kernels, which are invariant to translation of the inputs. These can be described as functions of $\tau = \vx - \vx'$,
\begin{equation}
k(\vx,\vx') = k(\vx - \vx').
\end{equation}
 
% ============================================================================================
\subsection{Spectral mixture kernels}\label{sec:sm}
% ============================================================================================
The class of flexible stationary kernels for GPs called Spectral Mixture (SM) kernels, was introduced in \cite{Wilson2013,Wilson2014}.
An SM kernel, here denoted by $k_\sm$, is derived through modeling its spectral density (the Fourier transform of a kernel) with a mixture of Gaussians.
Such modeling is possible because of Bochner's Theorem \cite{Bochner2016,Stein}, which states that the properties of a stationary kernel entirely depend on its spectral density: 

\begin{theorem}[Bochner]
A complex-valued function $k$ on $\bbR^P$ is the covariance function of a weakly stationary mean square continuous complex-valued random process on $\bbR^P$ if and only if it can be represented as
$k(\tau)= \int_{\bbR^P} e^{2\pi\imagi\vs \tra\tau}\psi(d\vs)$, 
where  $\psi$ is a positive finite measure, and $\imagi$ denotes the imaginary unit.
\end{theorem}

If $\psi$ has a density $\freq{k}(\vs)$, then $\freq{k}$  is called the spectral density or power spectrum of $k$, and $k$ and $\freq{k}$ are Fourier duals, that is: $$k(\tau)= \int \freq{k}(\vs)    e^{2\pi \imagi\vs \tra\tau} d\vs$$ and $$ \freq{k}(\vs)  = \int  k(\tau)  e^{2\pi \imagi \vs \tra\tau} d\tau.$$
The SM kernel $k_{\sm}$ is defined as the inverse Fourier transform (that is, the Fourier dual) of a mixture of $Q$ Gaussians in the frequency domain:
\begin{align}
\begin{split}
k_{\sm}(\tau)=&{\F}_{s\rightarrow \tau}^{-1}\bigg[\sum_{i=1}^Q{w_i}{{\freq{k}}_{{\sm}, i}}\bigg](\tau),
\end{split}\label{eq:sm-ft}
\end{align}
where ${\F}_{s\rightarrow \tau}^{-1}$ denotes the inverse Fourier transform operator from the frequency to the time domain, and $w_i$ is the weight of component $\freq{k}_{\sm,i}$. Here the spectral density $\freq{k}_{\sm, i}(\vs)$ of $i$-th component, is $$\freq{k}_{\sm, i}(\vs) = [\varphi_{\sm,i}(\vs) + \varphi_{\sm, i}(-\vs)]/2,$$
where ${\varphi}_{{\sm}, i}({\vs})={\N}({\vs};\vmu_{i},{\Var}_{i})$ is a scale-location Gaussian with parameters $\vmu_i,{\Var}_i$.
The symmetrization makes $\freq{k}(\vs)$ even, that is,  $\freq{k}(\vs) = \freq{k}(-\vs)$ for all $\vs$. So the Fourier transform of $\freq{k}$, that is our kernel, is real, since the Fourier transform of a real even function is real. Furthermore  $k$ is symmetric, since the Fourier transform of an even function is even (cf. e.g. \cite{hoffman1997introduction}).

By expanding equation \eqref{eq:sm-ft} we get 
\begin{subequations}
\begin{align}\label{eq:smp}
k_{\sm}(\tau)&= \sum_{i=1}^Q{w_i}k_{{\sm}, i}(\tau)\\
k_{{\sm}, i}(\tau)&={\cos\left(2\pi\tau\tra\vmu_{i}\right)}\prod_{p=1}^{P}{\exp\left(-2\pi^2\tau^{2}{\Var}_{i}^{(p)}\right)}
\end{align}
\end{subequations}
where 
$k_{{\sm}, i}$ is the $i$-th component, $P$ is the dimension of input, $w_i$, $\vmu_{i}=\left[\mu_{i}^{(1)},...,\mu_{i}^{(P)}\right]$, and ${\Var}_{i}=\text{diag}\left(\left[({\sigma_{i}^{2}})^{(1)},...,({\sigma_{i}^{2}})^{(P)}\right]\right)$ are weight, mean, and variance of the $i$-th component in the frequency domain, respectively. 

The inverse mean $1/\vmu_{i}$ of component $i$ is the  period, and the inverse standard deviation $1/\sqrt{{\Var}_{i}}$ the length scale, determining how quickly a component varies with the inputs. So the variance $({\sigma}_{i}^{2})^{(P)}$ can be thought of as an inverse length-scale, and $\mu_{i}^{(P)}$ as a frequency. 
The weight $w_i$ specifies the relative contribution of the $i$-th mixture component. 

In addition, by modeling the spectral density as a mixture of $Q$ univariate Laplacians, the kernel form of LKP \cite{Jang2017} can written as follows:
\begin{subequations}\label{eq:smp-lkp}
\begin{align}
k_{\lkp}(\tau)&= \sum_{i=1}^Q{w_i}k_{{\lkp}, i}(\tau)\\
k_{{\lkp}, i}(\tau)&={{\cos\left(2\pi\tau\tra\mu_{i}\right)} }\left({1+\frac{1}{2}\sigma^{2}_{i}{\tau}^{2}}\right)^{-1}
\end{align}
\end{subequations}

% ============================================================================================
\section{Skewed Laplace spectral mixture kernel}\label{sec:laplace}
% ============================================================================================

Any stationary covariance kernel can be approximated to arbitrary precision by an SM kernel, given enough  components of mixture in the spectral representation, because mixture of Gaussians are dense in the set of all distribution functions \cite{kostantinos2000gaussian}. So it would seem useless to introduce a new stationary kernel.
However, using a large number of components can lead to overfitting when there is not enough training data. Therefore, it makes sense to introduce and study new stationary kernels, that are robust to overfitting in the presence of a large number of components. An example of such a kernel is LKP.  However, as mentioned in the introduction, LKP does not model skewness characteristics of the spectral density beneficial for long-range forecasting.

Here we propose to overcome the limitations of the SM and LKP kernels
by using a mixture of Skewed Laplace (SL) distributions, which can better capture skewness of peaks of the spectral density and its heavy tail characteristic.
We consider the three-parameter ($\mu,\gamma, \sigma$) family of skewed Laplace distributions introduced in \cite{kotz2001asymmetric},
with density
\begin{align}\label{eq:laplace}
  \begin{split}
    \varphi(s;\mu,\gamma, \sigma)
    = & \frac{\sqrt{2}}{\sigma} \frac{\kappa}{1+\kappa^2} \begin{cases}
      \exp\left(-\frac{\sqrt{2}}{\sigma\kappa} (\mu -s)\right)
    &\text{if }s<\mu \\
      \exp\left(-\frac{\sqrt{2}\kappa}{\sigma}(s-\mu)\right)
    &\text{if }s\geq\mu
    \end{cases}
\end{split}
\end{align}
where $\kappa=\sqrt{2}\sigma / (\gamma+\sqrt{2\sigma^{2}+\gamma^{2}})$, 
$\gamma$ is
the skewness parameter,
$\mu$,  and ${\sigma^{2}}$ are the mean and variance of the distribution, respectively.
When $\gamma=0$ this distribution reduces to the standard non-skewed Laplace distribution.
The inverse Fourier transform of the $i$-th skewed Laplace density is \cite{fragiadakis2011goodness}
\begin{align}
\begin{split}
    \Finv[\varphi_{i}(s;\mu_{i},\gamma_{i}, \sigma_{i})](\tau) = &
    \frac{C_i(\tau)+\imagi\gamma_{i}{\tau}}{C_i(\tau)^2+\gamma_{i}^{2}{\tau}^{2}}{e^{\imagi\mu_{i}\tau}},
\end{split}
\label{eq:inv:sl}
\end{align}
where $C_{i}(\tau)=1+\frac{1}{2}\sigma^{2}_{i}{\tau}^{2}$.

We proceed to construct a kernel from the skewed Laplace density in the same way as described in the previous section:
1) symmetrize $\varphi_{i}(s;\mu_{i}, \gamma_{i}, \sigma_{i})$;
2) take the inverse Fourier transform of the resulting even real function; and
3) make a mixture of several such components.

First, by symmetrization we obtain the spectral density of the $i$-th component of the SL kernel:
\begin{align}\label{eq:freq_lsm}
\begin{split}
\freq{k}_{\lsm, i}(s) = &\frac{1}{2}\bigl( \varphi_{i}(s;\mu_{i},\gamma_{i}, \sigma_{i})+\varphi_{i}(-s;\mu_{i},\gamma_{i}, \sigma_{i})\bigr)
\end{split}
\end{align}

Note that the above transformation does not destroy the skewness of the mixture components (the peaks) of the spectral density: although by construction the spectral density is symmetric around zero, each of its peaks is skewed around its mean (with skew parameter $\gamma_i$).

Next, by applying \eqref{eq:inv:sl} to $\varphi_{i}(s;\mu_{i},\gamma_{i},\sigma_{i})$ and $\varphi_{i}(-s;\mu_{i},\gamma_{i}, \sigma_{i})$, and adding the resulting functions we obtain the time domain representation of the $i$-th component ${k}_{\lsm,{i}}$ of the SLSM kernel: 

\begin{align}\label{eq:lsm_component}
\begin{split}
{k}_{\lsm,{i}}(\tau)
=&\frac{1}{2}\bigg(\frac{C_{i}(\tau)+\imagi\gamma_{i}{\tau}}{C_{i}^{2}(\tau)+\gamma^{2}_{i}{\tau}^{2}}{\exp\left({\imagi\mu_{i}{\tau}}\right)}
+\frac{C_{i}(\tau)-\imagi\gamma_{i}{\tau}}{C_{i}^{2}(\tau)+\gamma^{2}_{i}{\tau}^{2}}{\exp\left(-{\imagi\mu_{i}{\tau}}\right)}\bigg)\\
=&\frac{C_{i}(\tau)\cos(\mu_{i}\tau)-\gamma_{i}{\tau}\sin(\mu_{i}{\tau})}{C_{i}^{2}(\tau)+\gamma^{2}_{i}{\tau}^{2}}.
\end{split}
\end{align}

Each component
${k}_{\lsm, i}(\tau)$ is positive semi-definite because its corresponding spectral density defined in \eqref{eq:freq_lsm}  is non-negative everywhere \cite{Bochner2016,Stein}. Also, each component is real because its spectral density is even \cite{Rasmussen2010}.

The denominator of ${k}_{\lsm, i}(\tau)$ is a skew function of the inverse distance and the numerator term is a periodical trigonometry function. $C_{i}(\tau)$ is an inverse Cauchy function (the inverse Fourier transform of the zero positioned non-skewed Laplace) \cite{Jang2017}. The Cauchy function has slower decay over the distance $\tau$ than the exponential function used in SM based kernels.

Finally, the Skewed Laplace Spectral Mixture (SLSM) kernel ${k}_{\lsm}$ corresponds to a mixture of $Q$  skewed Laplace components, and is defined as
\begin{align}
\label{eq:slsm}
\begin{split}
k_{\lsm}(\tau)
&=
{\F}_{s\rightarrow \tau}^{-1}\bigg[\sum_{i=1}^Q{w_i}{{\freq{k}}_{{\lsm}, i}}\bigg](\tau) \\
&=\sum_{i=1}^{Q}w_{i}\frac{C_{i}(\tau)\cos\left({{\mu}_{i}}{\tau}\right)-\gamma_i\tau\sin\left({{\mu}_{i}}{\tau}\right)}{C_{i}^{2}(\tau)+ {\gamma^{2}_{i}}\tau^{2}}.
\end{split}
\end{align}
 
The Skewed Laplace distribution has a natural extension to higher dimensions, described, e.g., in \cite{kotz2001asymmetric,Kozubowski2000,visk2009parameter}. 
Therefore our kernel can be directly extended to the multivariate setting as follows: 

\begin{align}\label{eq:lsm_multi}
\begin{split}
{k}_{\lsm}(\vtau)
&=\sum_{i=1}^{Q}{w_{i}}\frac{C_{i}(\vtau)\cos\left({{\vtau}^{\top}}{{\vmu }_{i}}\right)-({\vtau^{\top}\vgamma_i})\sin\left({{\vtau}^{\top}}{{\vmu }_{i}}\right)}{C_{i}^{2}(\vtau)+ ({\vtau^{\top}\vgamma_i})^{2}},
\end{split}
\end{align}
where  $C_i(\vtau) = 1 + \frac{1}{2} \vtau^\top \Sigma_i\vtau$. Here  $\Sigma_i$ denotes the covariance matrix of component $i$, and ${\vmu }_{i}$ is the vector of means of this component.

\subsection{Comparison with other kernels}

Figure~\ref{fig:lsm-spec} provides a visual comparison of the SM, LKP and SLSM distributions used for fitting the empirical spectral density.  The differences among SLSM, LKP and SM are illustrated in terms of covariance, spectral density, and sampling functions.  The subplots in the second row of Figure~\ref{fig:lsm-spec} show random functions  drawn from a GP with SM, LKP, and SLSM kernel, respectively. The sampled function values were obtained using $500$ equally-spaced discrete points.

The inverse FT of the SM  spectral density is an exponential function, so in the time domain we have an exponential decay of covariance. When we use LKP to fit the empirical spectral density and apply the inverse FT  we get a Cauchy function, which decays in a much slower way  than the exponential one.
Using SLSM we extend the tails of the Laplace distribution (by means of the $\gamma$ term in equation \ref{eq:laplace}), which in the time domain yields a further reduction of the decay rate of the covariance. This phenomenon is illustrated in Figure~\ref{fig:lsm-spec}, which shows longer range covariance of SLSM (middle of right plots).  
The main characteristics of SM, LKP and SLSM can be summarized as follows: 
\begin{itemize}
    \item SM: multivariate Gaussian, dense, non-skewed peaks, short-tailed, smooth, symmetry of each peak, exponential decay of covariance;
    \item LKP: univariate Laplacian, sparse, non-skewed peaks, heavy-tailed,  non-smooth, symmetry of each peak, fast decay of covariance;
    \item SLSM: multivariate skewed Laplacians, skewed peaks, one side more heavily-tailed, non-smooth, slower decay of covariance.
\end{itemize}

Figure \ref{fig:lsm-init-spec} shows an example where SLSM provides a better fit of the empirical spectral density than SM and LKP.
SLSM extends the LKP kernel (apart from the way inference is performed), which can be obtained by removing the skewness term, i.e., by setting the $\gamma_{i}$'s to $0$. More precisely, LKP as defined in \cite{Jang2017}, uses Reversed-Jump MCMC (RJ-MCMC) to perform inference. While, SLSM uses LBFGS implemented in GPML toolbox to perform inference. 

Also,  SLSM is related to the Rational Quadratic (RQ) kernel,
\begin{align}\label{eq:rq}
\begin{split}
{k}_{\text{RQ}}(\tau)&
=\theta_{f}\left(1 +\frac{\tau^{2}}{2\alpha\ell^{2}}\right)^{-\alpha}
\end{split}
\end{align} 
The RQ kernel can be interpreted as a scale mixture (an infinite sum) of squared scale mixture SE kernels with different characteristic length-scales \cite{Rasmussen2006}.
The RQ kernel is considered  the most general representation defining a valid isotropic covariance function in $\mathbb{R}^{P}$ \cite{Wilson2014,Stein}. 
From Equations~\eqref{eq:lsm_component} and \eqref{eq:rq} it follows that a non-skewed SLSM component ${k}_{\lsm, i}(\tau)$ is the product of a RQ kernel with $\alpha=1$, $\theta_{f}=w_{i}$,  $\ell^{-2}=\sigma^{2}_{i}$ and a cosine kernel. 
Thus the RQ kernel with $\alpha=1$ can be viewed as modeling the spectral density, as one Laplace distribution at zero mean position in the frequency domain. That is,  the RQ kernel can be viewed as part of a component of a SLSM kernel.

SLSM component can be approximated at arbitrary precision with a mixture of Gaussians using sufficiently many components. Thus a SM kernel can approximate the SLSM kernel.
However, as observed e.g. by \cite{samo2015generalized}, although mixture of Gaussian distributions can be used to approximate any stationary distribution, a large number of SM components might be required to account for the lower degrees of smoothness in the data. This would affect inference, which would become more expensive and be more vulnerable to the presence of local optima. In Figure~\ref{fig:lsm-spec}  SM, LKP, and SLSM with 3 components are illustrated in terms of spectral densities, covariance, and sampling paths.

%----------------------------------------------------------------------------------------
\section{SLSM's hyper-parameter initialization, automatic pruning of components, and scalable inference}\label{sec:auto-init}

Kernels hyper-parameters are inferred by optimizing the negative log marginal likelihood (NLML), which for kernels such as SM and SLSM, amounts to solve a non-convex optimization problem. The initialization of these hyper-parameters has a direct impact on the ability of the optimization process to find a good local optimum of the NLML. Below, we describe the initialization procedure used in our experiments.
Another parameter of spectral kernels to be set is the number of kernel components. A way to circumvent this problem is to automatically prune irrelevant components.  In LKP, automatic pruning of components is part of the method, because of the sparsity-inducing property of the L\'evy priors (see \cite{Jang2017}, supplementary material).  For SLSM and SM kernels, we propose a method for automatic pruning kernel components. In order to perform experiments with large datasets, scalable inference is needed, because in GPs exact inference is computationally expensive. Therefore, for the large-scale experiment in Section \ref{large-dataset} we use the scalable inference procedure described in this section. 

\subsection{Hyper-parameter initialization}\label{sec:init}
%----------------------------------------------------------------------------------------

In general, SM-based kernels are particularly sensitive to the initialization of their hyper-parameters,
and the SLSM kernel shares this initialization problem.   
Here we apply an initialization strategy that has been shown to be effective in previous works \cite{Wilson2013,herlands2016scalable}.
These works suggest that non-smooth peaks of the empirical spectral densities are near the true frequencies and use a Gaussian Mixture Model to fit the empirical spectral density, which is then used to initialize the hyper-parameters of an SM kernel.
We make use of this result  and initialize the hyper-parameters of the SLSM kernel using a Laplace Mixture Model,
\begin{align}
    p_{\text{LMM}}({{\Theta |\vs}})=\sum _{i=1}^{Q}{\tilde {w}_{i}}{\phi_{i}}(\tilde{\mu}_{i},{\tilde{\Sigma}_{i}})
\end{align}
 to fit the empirical spectral density in order to get $Q$ cluster centers, where $Q$ is the number of  components.
 We use the Expectation Maximization algorithm \cite{Moon1997The} to estimate the parameters of this mixture model. 
 The resulting estimates  $\tilde{w}_{i}$, $\tilde{\mu}_{i}$, and $\tilde{\Sigma}_{i}$ are used as initial hyper-parameters values of the SLSM kernel.
 The skewness parameter $\gamma_{i}$ is initialized randomly between -1 and 1.
We use the above initialization strategies in all our experiments on real-world datasets.

%=======================================================
\subsection{Automatic pruning of kernel components}\label{sec:prune}
%=======================================================
In order to automatically select the number of components, we adopt a method originally introduced for pruning the weights of a neural network, based on the Lottery Ticket Hypothesis (LTH): dense, randomly-initialized, feed-forward networks contain subnetworks (winning tickets) that - when trained in isolation - reach test accuracy comparable to the original network in a similar number of iterations \cite{DBLP:conf/iclr/FrankleC19}.  The above mentioned method trains a neural network and prunes a percentage of small weights; the process is repeated in a recursive fashion for each subsequent network. At each iteration, each unpruned weight value is reset to its initial value before training. 

In our context, we want to prune Laplace kernel components as part of training the model. This means that the components that we prune are the Laplace kernel components that make up equation \eqref{eq:slsm}. 
Note that kernel components should not be confused with the sines and cosines terms of the Fourier decomposition of the empirical spectral density, which are often also referred to as components. 
Each component of the kernel covers many frequencies in the spectral density.

Specifically, in our case components
with magnitude of the weight $w_i$ smaller than $1$ are pruned. The choice of this threshold is heuristically motivated: since the energy of spectral density is very big, if a weight is smaller than 1, then the contribution to its component is very small and can be ignored. Figure~\ref{fig:sun-spot-hist} illustrate this phenomenon, and shows the weights of components of kernels of a GP trained on the sunspot data (described in Section~\ref{sec:exp_sun}) using $Q=100$ components.  Weights of many components (almost $65\%$) of SLSM are smaller than 1, while for SM and LKP  there are less than $30\%$ components with weights smaller than 1.

In our experiments we use the following LTH algorithm to prune components of SM and SLSM kernels for GP:
\begin{enumerate}
    \item Initialize a GP with $Q$ components.
    \item Train the GP for $100$ iterations using the LBFGS algorithm (see description below).
    \item Prune component with weight smaller than $1$. %, $w_{i}^{(j)}$ in $\Theta^{(j)}$.
    \item Set weights of unpruned components to their initial value and re-train the GP. 
\end{enumerate}
We train the GP model by means of the Limited-memory Broyden-Fletcher-Goldfarb-Shanno (LBFGS) algorithm with 100 iterations.
The computational cost of this pruning procedure depends on the number of iterations performed. In practice, two pruning rounds are sufficient.

Our LTH pruning algorithm is applicable to SM and SLSM. We also consider the specific variant of SLSM obtained by setting  $\gamma=0$ (we call this variant SLSM($\gamma=0$)). This variant is strongly related with the LKP kernel (see Section \ref{sec:exp}).
As already mentioned, in the LKP method,  the L\'evy process introduces a sparsity-inducing prior over mixture components, allowing
automatic pruning of components.

\subsection{Scalable inference}
For GPs, exact inference is prohibitively slow for more than a few thousand of points because of its $\compO(n^{3})$ computational complexity and $\compO(n^{2})$ memory complexity  \cite{Rasmussen2006,Rasmussen2010,Quinonero-Candela2005,Chalupka2013} in the Cholesky decomposition of covariance matrix. 
The most computationally expensive steps are inverting the covariance matrix and computing the determinant of the covariance (i.e. $(K+\sigma^{2}_{n}I)$), see Eq.~\eqref{eq:gp_pred} and Eq.~\eqref{eq:nlml}. These problems have been addressed by covariance matrix approximation or inference approximation. 

In our experiment with big time series of length $n\geq 10000$, we adopt the robust Bayesian Committee Machine (rBCM) of distributed GPs \cite{deisenroth2015distributed} framework to approximate the covariance matrix structures for scalable inference. rBCM is a scalable product-of-experts model for large-scale GP regression, which recursively distributes computations to $M$ independent local computational units and, recombine them to form an overall result. 
Due to the independence between local GP models, the marginal likelihood (see Eq. (\ref{eq:nlml})) of a full GP using rBCM can be factorized into a product
form of $M$ marginal likelihoods of local GP models, thus performing scalable inference of the full GP model. 

Using rBCM, we can train a GP model with SLSM kernel as follows:
\begin{align}
p(\vy|X, \Theta) \approx\prod_{i=1}^{M}p^{(i)}(\vy^{(i)}|X^{(i)}, \Theta)
\end{align}
where $M$ is the number of local GPs and $p^{(i)}(\vy^{(i)}|X^{(i)}, \Theta)$ is the marginal likelihood of the $i$-th local GP using the $i$-th subset $\{X^{(i)}, \vy^{(i)}\}$ of dataset $\{X, \vy\}$. 
The predictive probability is computed as the product of the  predictive probabilities of the independent local GPs, as follows:
\begin{align}
p(f_{*}|\vx_{*}, \vy, X) \approx\prod_{i=1}^{M}p^{\beta_i}(f_{*}|\vx_{*}, \vy^{(i)}, X^{(i)}).
\end{align}
where $\beta_i$ is the weight of $i$-th local GP.

%===============================================================================
\section{Experiments}\label{sec:exp}
%===============================================================================

\begin{figure*}[ht!]
    \renewcommand{\tabcolsep}{0mm}
    \def\incpic#1{\includegraphics[width=0.33\columnwidth]{#1}}
    \def\incpicslsm#1{\includegraphics[width=0.34\columnwidth]{#1}}
    \centering
    \begin{tabular}{@{}*{3}{c}}
        \incpic{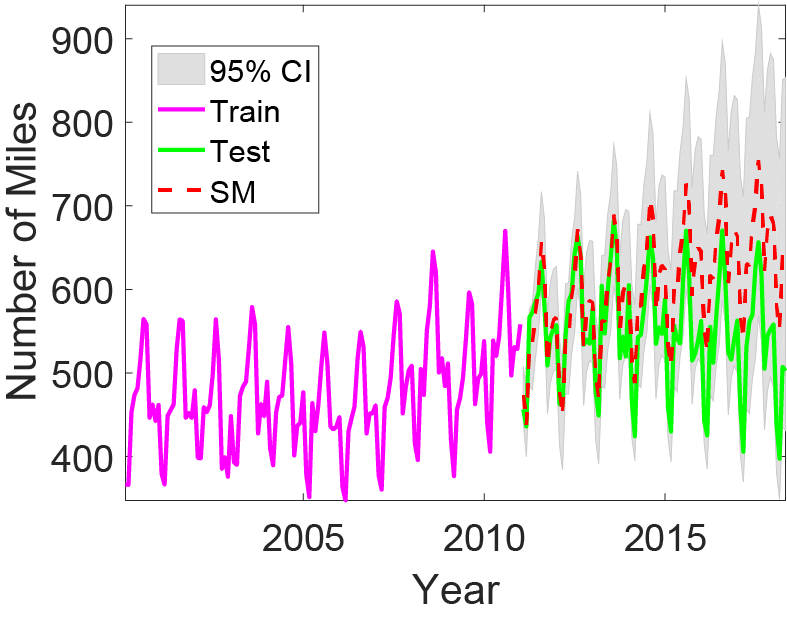} &
        \incpic{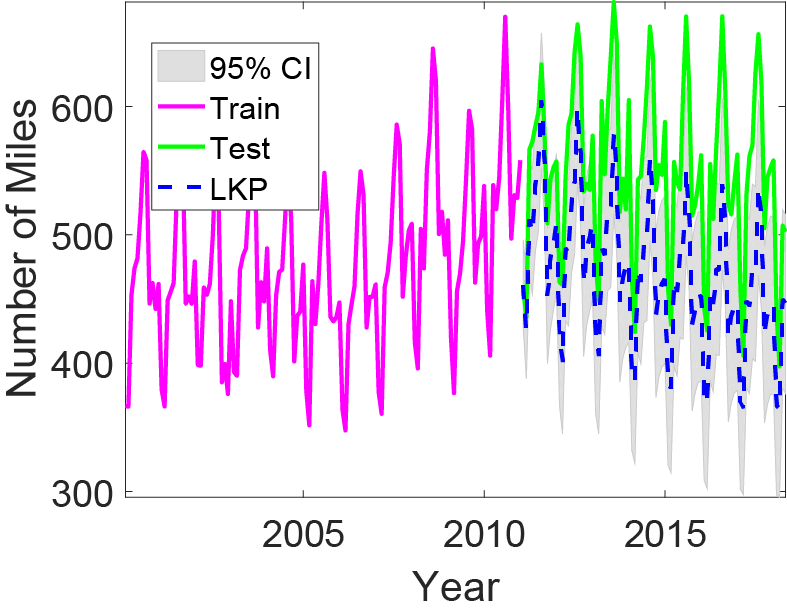} & 
        \incpicslsm{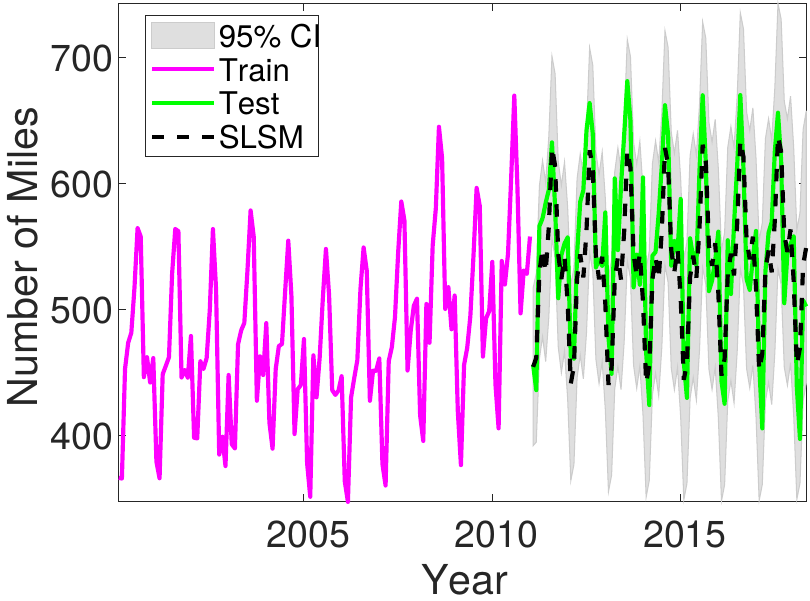} \\
        \incpic{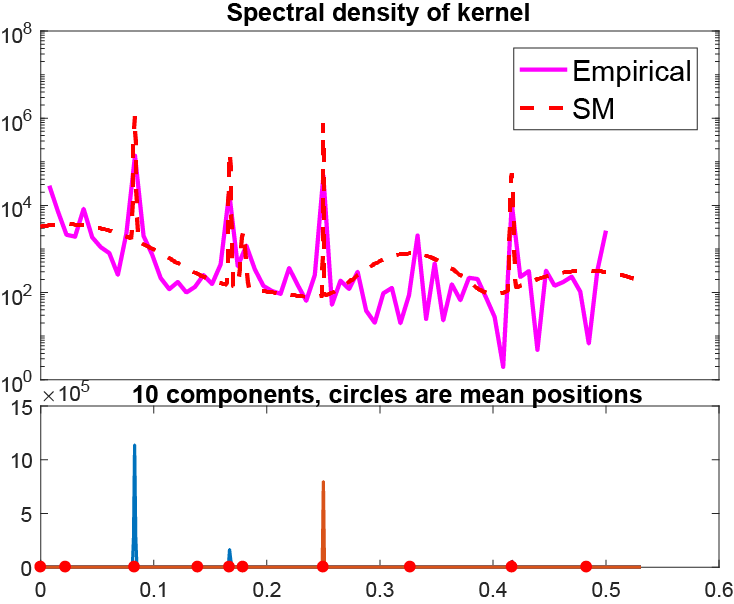} & 
        \incpic{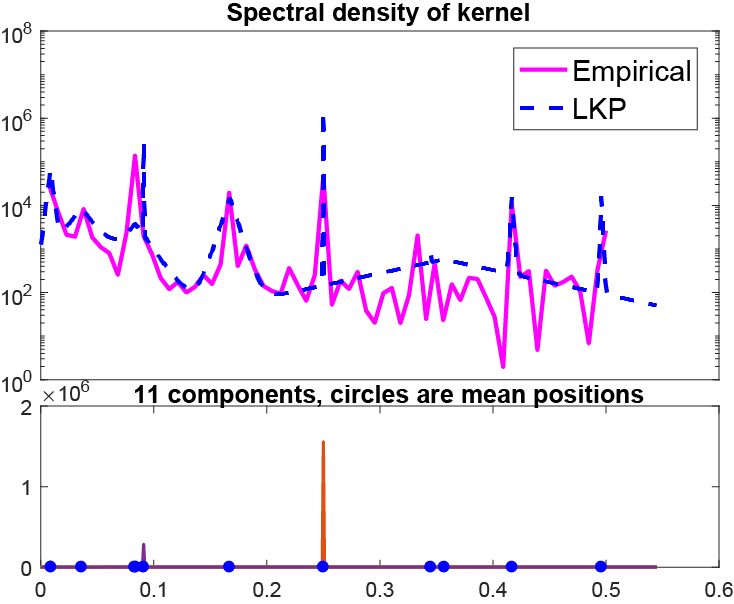} & 
        \incpicslsm{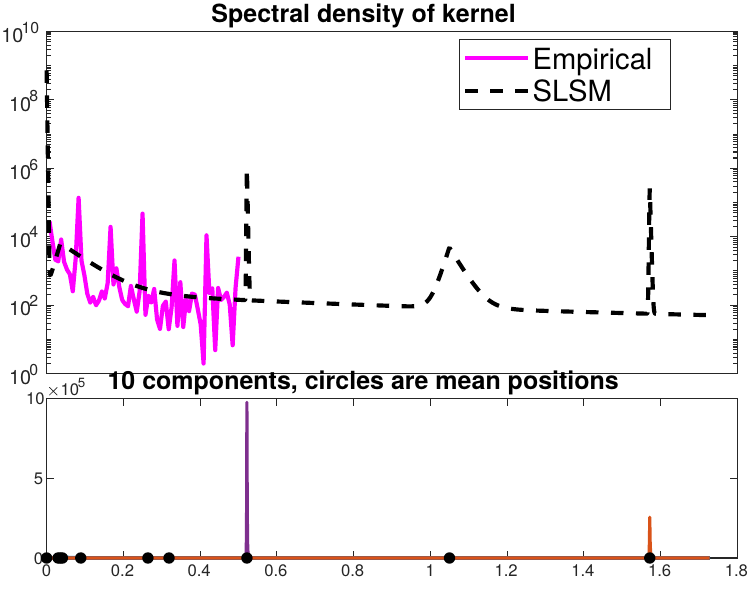} \\
        {(a) SM} & {(b) LKP (L\'evy prior)} & {(c) SLSM}\\
    \end{tabular}
    \caption{Extrapolation results on the Rail Passenger Miles dataset. Bottom plots: spectral densities of optimized kernels, together with  the empirical spectral density of the training data. Note that the better fitting of SM and LKP on the empirical spectral densities (training data) may lead to overfitting and negatively affect extrapolation performance.}
    \label{fig:rail}
\end{figure*}

\begin{figure*}[ht!]
    \renewcommand{\tabcolsep}{0mm}
    \def\incpic#1{\includegraphics[width=0.335\columnwidth]{#1}}
    \centering
    \begin{tabular}{*{3}{c}}     
        \incpic{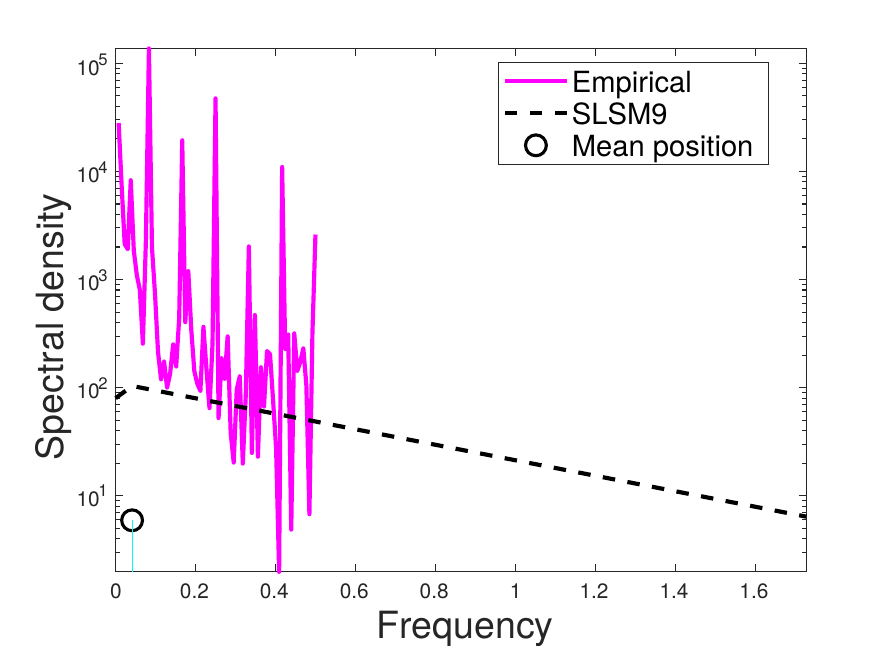} 
        & \incpic{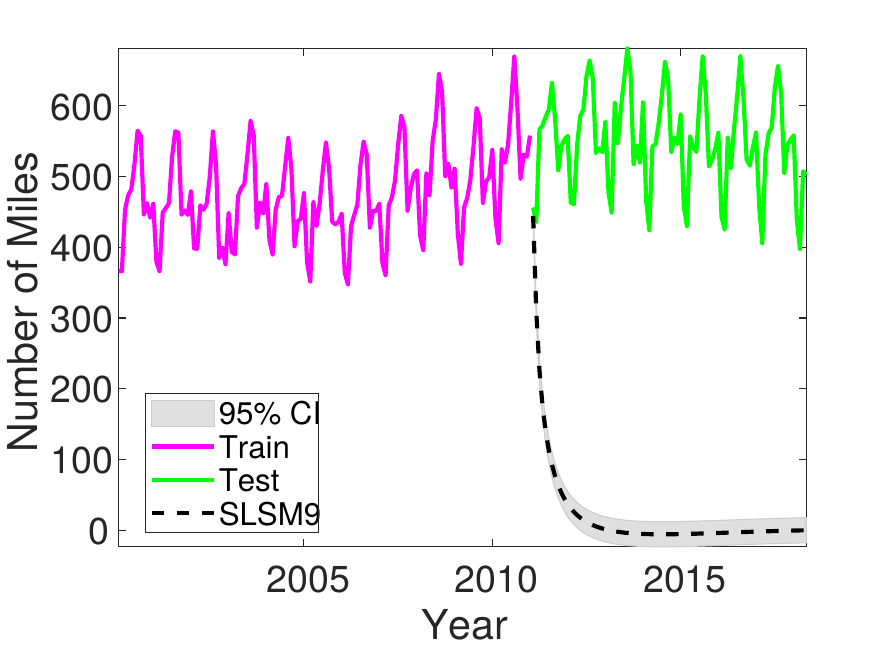} 
        & \incpic{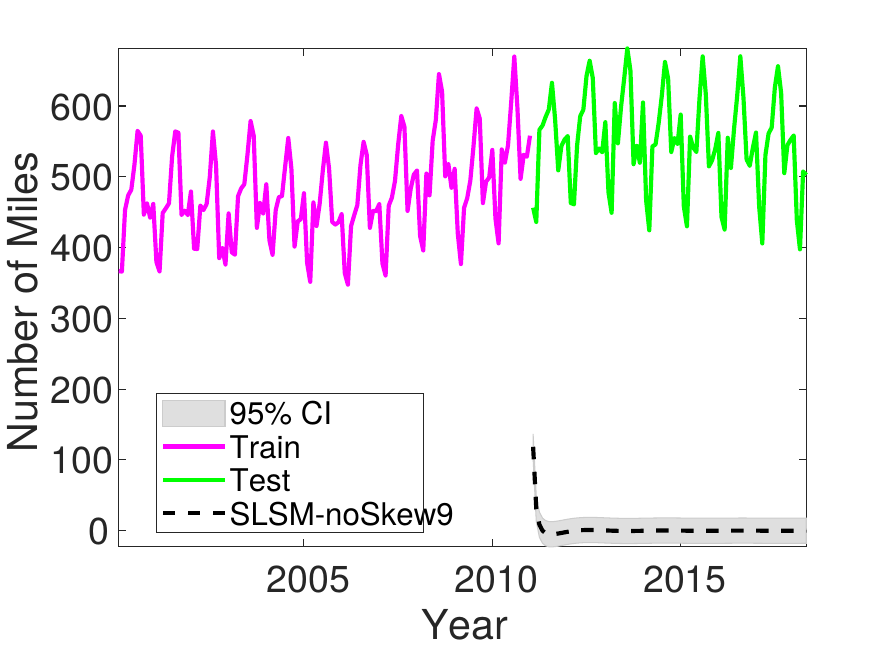} \\
        {(a) Skewed $\freq{k}_{\lsm,9}(s)$} 
        & {(b) $f_{9}\sim{}\gp(0, {K}_{\lsm,9}(\tau))$}  
        & {(c) $f_{9}\sim{}\gp(0, {K}_{\lsm,9}(\tau, \gamma=0))$}
    \end{tabular}
    \caption{
    An instance of skewed peak ($\mu_{9}=0.04, \gamma_{9}=0.45$) learnt by SLSM, its posterior extrapolation, and non-skewed ($\gamma=0$) posterior extrapolation. With skewness, $f_{9}\sim{}\gp(0, {K}_{\lsm,9}(\tau))$ can keep trend for a longer time than $f_{9}\sim{}\gp(0, {K}_{\lsm,9}(\tau, \gamma=0))$. }
    \label{fig:rail-component}
\end{figure*}

We assess the performance of SLSM and of the following popular stationary kernels available in the GPML toolbox \cite{Rasmussen2006,Rasmussen2010}:
linear with bias (LIN), squared exponential (SE), piecewise polynomial (Poly), periodic (PER), rational quadratic (RQ), Mat\'ern 5/2 (MA), Gabor, neural network (NN), additive kernel based on SE using unary and pairwise interactions (ADD), as well as SM and LKP kernels. Also, we assess the performance of the LKP kernel as an instance of SLSM with $\gamma=0$, we denote this variant by SLSM($\gamma=0$). LKP and SLSM($\gamma=0$)  have the same kernel structure, but inference is performed using different optimization methods: LKP uses reversible jump Markov chain Monte Carlo (RJ-MCMC), while SLSM($\gamma=0$) uses the standard LBFGS method based on gradient-based optimization.

Also, we consider popular traditional forecasting methods, namely autoregressive integrated moving average (ARIMA), automatic ARIMA (Auto-ARIMA) and the error, trend, seasonality (ETS) model with Holt-Winters seasonal method. These methods are different from GP models and have been used for forecasting time series with trend and seasonal components. We use the implementation of the ARIMA, Auto-ARIMA, and ETS in the ECOTOOL toolbox, with default parameters values \cite{pedregal2019time}.

For SM, LKP, and SLSM, we consider $Q=10$ components, which are then automatically pruned.    The choice of Q=10 is heuristic, as e.g. in  \cite{Wilson2013}. The underlying assumption is that each component corresponds to a distribution in the frequency domain, which covers a much wider range of frequencies than a sine or cosine located at a single frequency position. Therefore a mixture of $10$ components located at different positions can cover the structure of the underlying spectral density reasonably well, that is, the underlying peaks (determined by the weight), frequency range (determined by the variance), skewness,  peak positions (determined by the mean) of spectral density in the learned model other than empirical spectral density).

For SM we initialize the hyper-parameters using a Gaussian Mixture Model $p_{\text{GMM}}({{\Theta |\vs}})=\sum _{i=1}^{Q}{\tilde {w}_{i}}{\N}(\tilde{\mu}_{i},{\tilde{\Sigma}_{i}})$. 
For LKP we use a non-skewed Laplace Mixture Model.
All other kernels in our comparison use the standard initialization in GPML toolbox.

The LKP method involves a L\'evy process prior and RJ-MCMC to prune extraneous components and to infer the posterior distribution, respectively, as described in \cite{Jang2017}.
For all other kernels, we use exact inference together with LBFGS for optimizing the kernel hyper-parameters.   

For SM and SLSM, we also report results obtained by pruning extraneous components using the LTH algorithm described in Section~\ref{sec:prune}. Also, plots of the spectral density, specific skewness values, instance of skewed component, and experiments with varying values of $Q$ in SLSM, LKP, SM are considered to provide a deeper insight into their performance.

In all experiments we use the following evaluation metrics: mean squared error (MSE)
${\mathrm {MSE} ={\frac{1}{n}{\sum _{i=1}^{n}\left(y_{i}-\tilde{y}_{i}\right)^{2}}}}$, mean absolute error (MAE) ${\mathrm {MAE} ={\frac{1}{n}{\sum _{i=1}^{n}\left|y_{i}-\tilde{y}_{i}\right|}}}$, where $\tilde{y}_{i}$ denotes the predicted value, and NLML (computed over the training data). Results are average  over $10$ independent runs. Standard deviations are also reported.

We consider many forecasting tasks with real-world datasets: 
rail miles, monthly electricity, mean sunspot, computer hardware, the monthly M3-competition dataset with more than 1000 time series, and the large pole telecom dataset.  
Source code of the proposed SLSM kernel is available at \url{https://github.com/ck2019ML/GPs-for-long-range-forecatsing}.

\begin{figure*}[th!]
    \renewcommand{\tabcolsep}{0mm}
    \def\incpic#1{\includegraphics[width=0.33\columnwidth]{#1}}
    \centering
    \begin{tabular}{*{3}{c}}
        \incpic{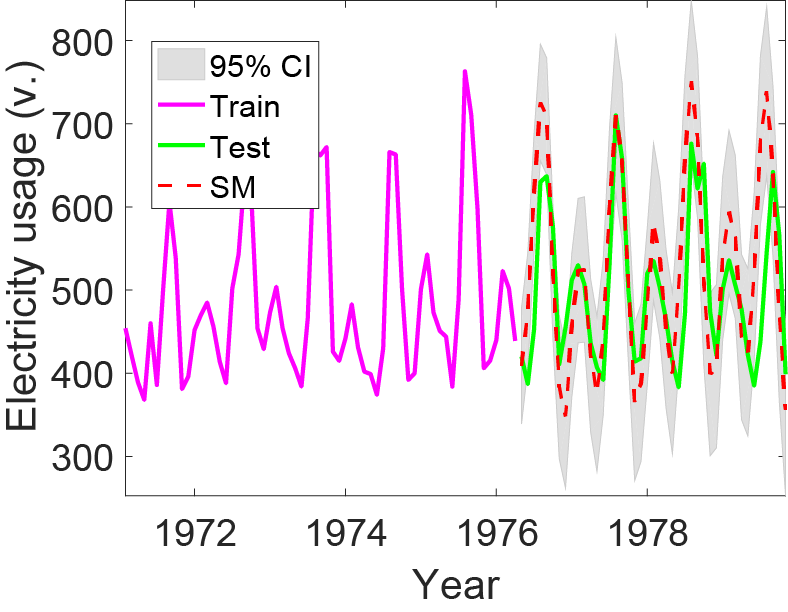} &
        \incpic{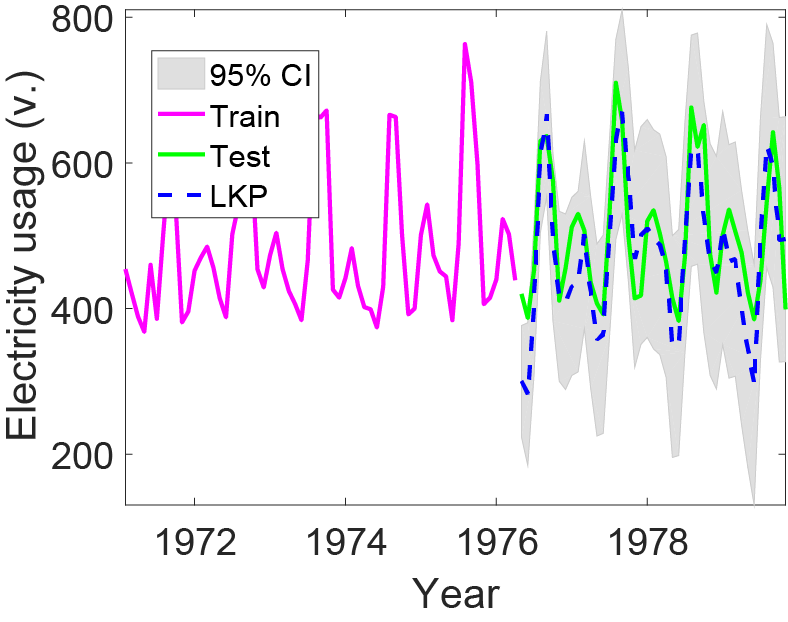} &
        \incpic{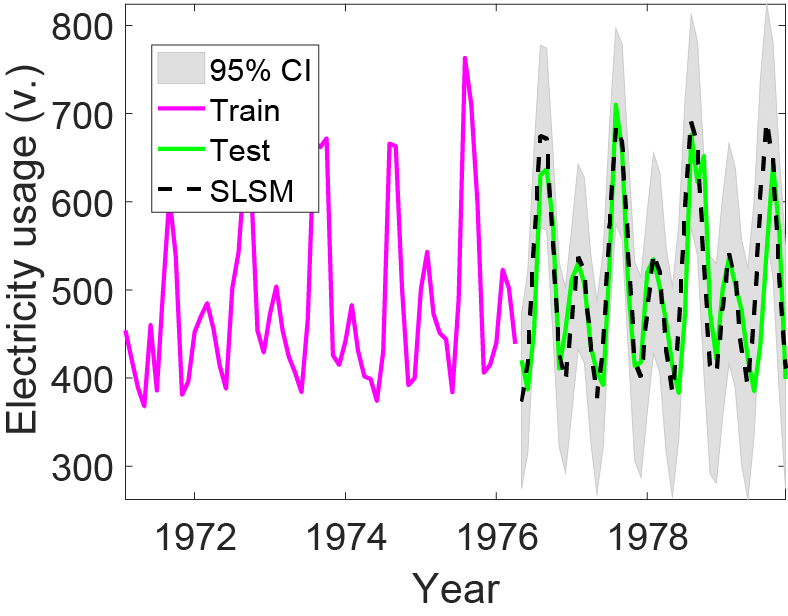} \\
        {(a) Extrapolation of SM} &
        {(b) Extrapolation of LKP} & {(c) Extrapolation of SLSM} \\
        \incpic{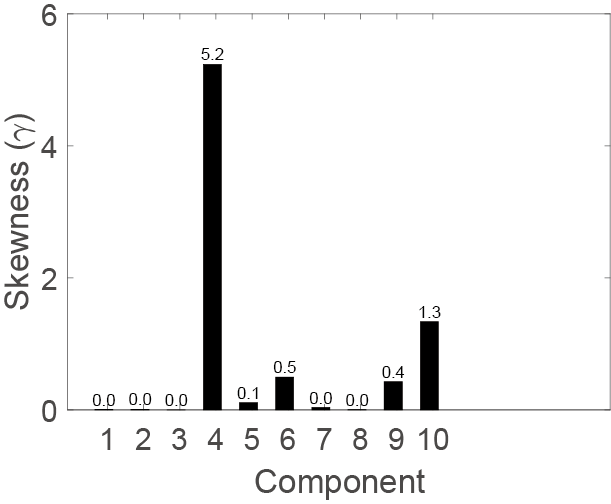} 
        & \incpic{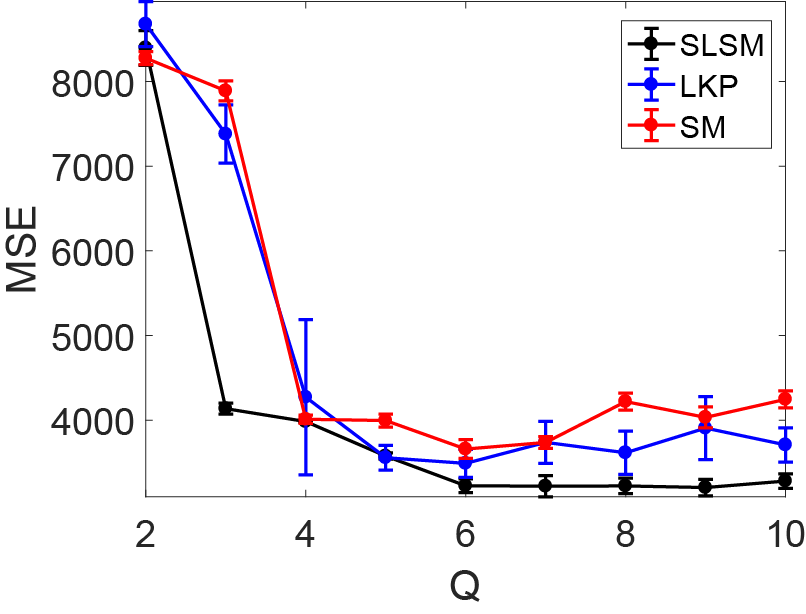} 
        & \incpic{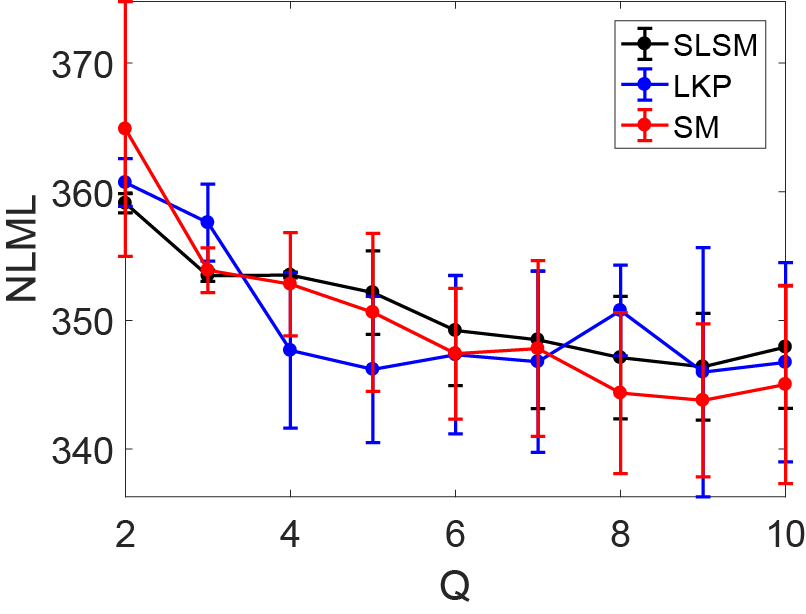} \\
        {(d) Skewness of SLSM} & {(e) MSE of varing $Q$} & {(f) NLML of varing $Q$}
    \end{tabular}
    \caption{
    Extrapolation using the SM kernel (a),
    LKP kernel using L\'evy prior (b),
    and SLSM kernel (c) on the monthly electricity usage dataset.
    The solid magenta line shows the training data, while the dashed line is the mean prediction by a GP.
    Plots (d), (e) and (f) show the skewness of the optimized SLSM kernels, together with MSE and NLML of varying Q. 
    All plots for the other experiments use the same notation.}
    \label{fig:elec}
\end{figure*}

%----------------------------------------------------------------------------------------
\subsection{Long-term forecasting of rail passenger miles: comparing spectral densities of SM, LKP and SLSM} 
\label{sec:exp_rail}

In this experiment, we illustrate  on a real-life long-term forecasting problem the differences between SM, LKP and SLSM kernels in terms of their spectral densities. 
We consider the rail miles dataset \footnote{https://fred.stlouisfed.org/series/RAILPM}.
This dataset consists of 220 values covering the period from Jan. 2000 to Apr. 2018.
We use $Q=10$ components for the SM, LKP and SMLS kernels
The top plots in Figure~\ref{fig:rail}  show extrapolation results.
For test points far from the last (training) observation, the SLSM kernel achieves best extrapolation performance.
The bottom part of Figure~\ref{fig:rail} shows that the spectral density of the SLSM kernel has components with a higher variance than that of the other kernels as well as components with higher mean.

In Figure~\ref{fig:rail} we can observe that better fit on empirical spectral density of SM and LKP lead to limited generalization in terms of extrapolation. This indicates that the fitting on empirical spectral density is a poor proxy for the generalization error of testing, since the model may be overfitted, leading to low training error but poor generalization (testing) performance. In contrast, the optimized SLSM kernel looks different from the empirical spectral density. We believe that this is the result of the extra flexibility that skewness provides, which allows the optimizer to find a better local optimum for the kernel hyper-parameters. Which in turn leads to better generalization, since it can capture patterns not present in the empirical spectral density.

Looking in more detail, we can see that the optimized SLSM kernel has two components with large mean, $\mu_{5}=1.57$ and $\mu_{5}=1.05$, which may respectively correspond to smaller periods of about half month and one month, in line with half and one monthly variations that appear in the dataset.  In SM and LKP, their spectral densities miss many of  variations at higher frequencies, meaning they cannot learn structures located at higher frequencies.
In Figure \ref{fig:rail-component} we show spectral density, extrapolation, and non-skewed extrapolation of the $9$th skewed SLSM component in SLSM.  
Subplots (b) and (c) in Figure \ref{fig:rail-component} show that $f_{9}$ reduces to zero fast if we just remove its skewness and keep the rest hyper-parameters unchanged, demonstrating that skewness extends covariance range in GP model, which can correlate test points far away from training points and thus is beneficial for long range extrapolation. 

The specific values of weights $w$, means $\mu$, variances $\var$, and skewness $\gamma$ in SLSM  determine the contribution, periodicity, length-scale, and long-range dependency of each SLSM component, respectively. 

Overall, results indicate that GP with SLSM kernel achieves best performance 
 in terms of MSE (see Table~\ref{tab:MSE}) and MAE (see Table~\ref{tab:MAE}), while its  NLML is bigger than that of SM (see Table~\ref{tab:NLML}). This may seem surprising. However, note that the computation of NLML is based on training data (see Eq. (\ref{eq:nlml})). Therefore a lower NLML value usually indicates a better fitting of the training data, but does not guarantee better generalization.
Also, note that NLML is the sum of two terms (and a constant term that is ignored): a model fit and a complexity penalty term. The first term is the data fit term which is maximized when the data fits the model very well. The second term is a penalty on the complexity of the model, i.e. the smoother the better. When optimizing NLML finds a balance between the two and this changes with the data observed.

%----------------------------------------------------------------------------------------
\subsection{Long-term forecasting of electricity consumption: varying the number of kernel components}\label{sec:exp_elec}
%----------------------------------------------------------------------------------------

In this experiment we investigate comparatively the performance of SM, LKP and SLSM  when varying the number of components in a small range (2-10), using real-life data for long-term monthly electricity forecasting. Electricity forecasting is a problem of practical relevance  for electric  power  system  planning,  tariff  regulation  and  energy  trading  \cite{pessanha2015forecasting}.
The dataset includes monthly residential electricity usage in Iowa city from Jan. 1971 to Oct. 1979. There are a total 106 time points, of which we use the first 60$\%$ for training and the rest for testing.

For SLSM, LKP, and SM we use $Q=10$ components.
Figure~\ref{fig:elec}, subplots (a), (b), and (c), show that SLSM has a better extrapolation performance than SM and LKP, in particular for small peaks corresponding to the winter months. 
Subplot (d) shows the importance of skewness in SLSM, with a component having a relatively very high skewness value. 

On this data, LKP selects automatically $11$ components. 
Figure~\ref{fig:elec} (subplots (e) and (f))  shows the performance of  SLSM, SLSM($\gamma=0$), and SM with varying number of components. 
Using the same value of $Q$ for all kernels seems unfair, because SLSM has $4Q$ parameters while  SM and LKP have $3Q$ parameters. Comparing performance at same value of the number of parameters would mean to increase the value of $Q$ for SM and LKP.   However, using a bigger $Q$ does not necessarily yield better performance. For instance, results indicate that SLSM with $Q=6$ is still better than  SM and LKP with $Q=10$. 
Results in terms of MSE, MAE, and NLML are given in Tables~\ref{tab:MSE}, Tables~\ref{tab:MAE} and~\ref{tab:NLML}, respectively. 

\begin{figure*}[ht!]
    \renewcommand{\tabcolsep}{1mm}
    \def\incpic#1{\includegraphics[width=0.32\columnwidth]{#1}}
    \centering
    \begin{tabular}{*{3}{c}}
        \incpic{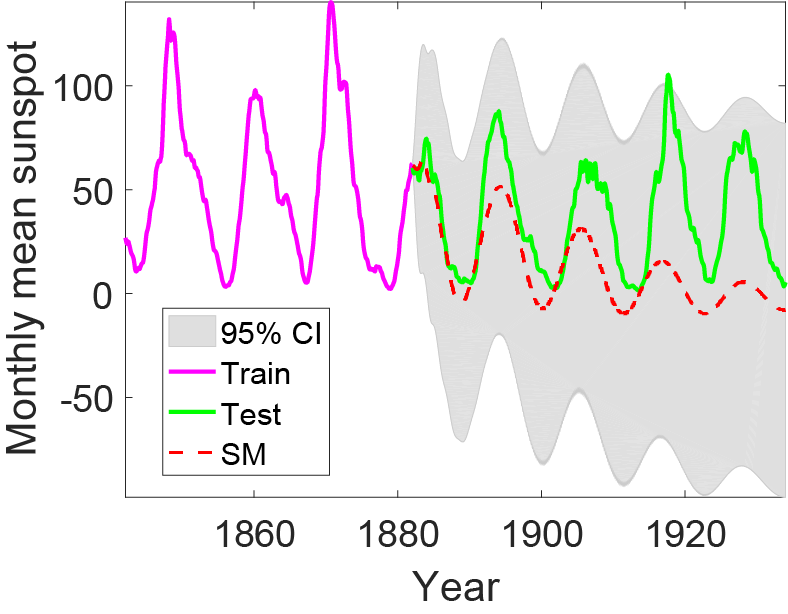} &
        \incpic{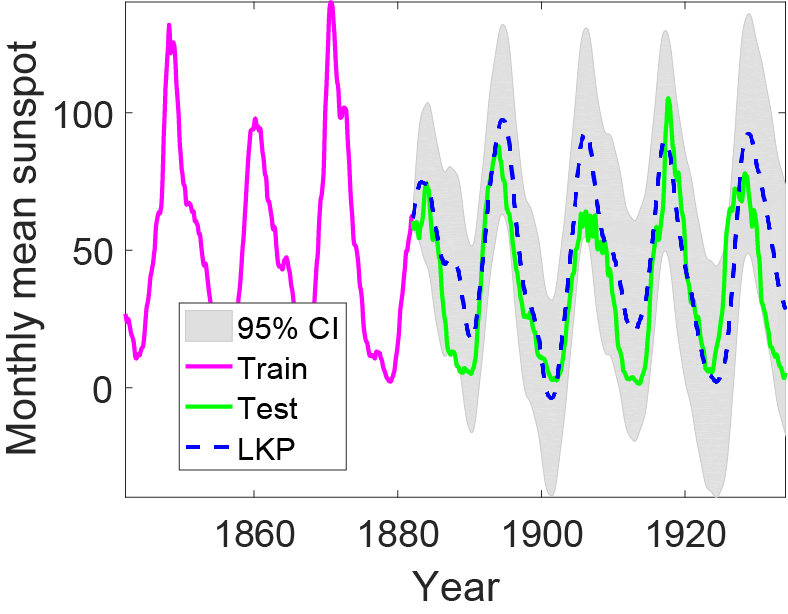} &
        \incpic{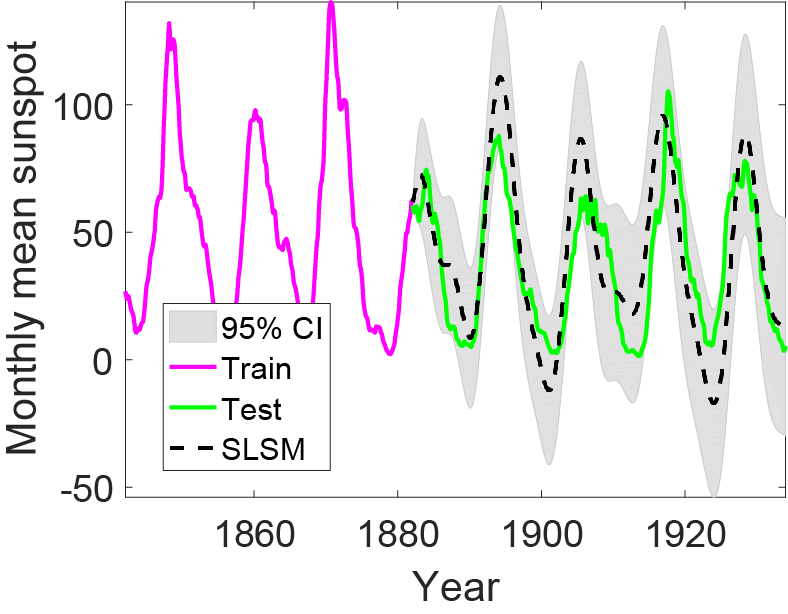}
        \\
        \incpic{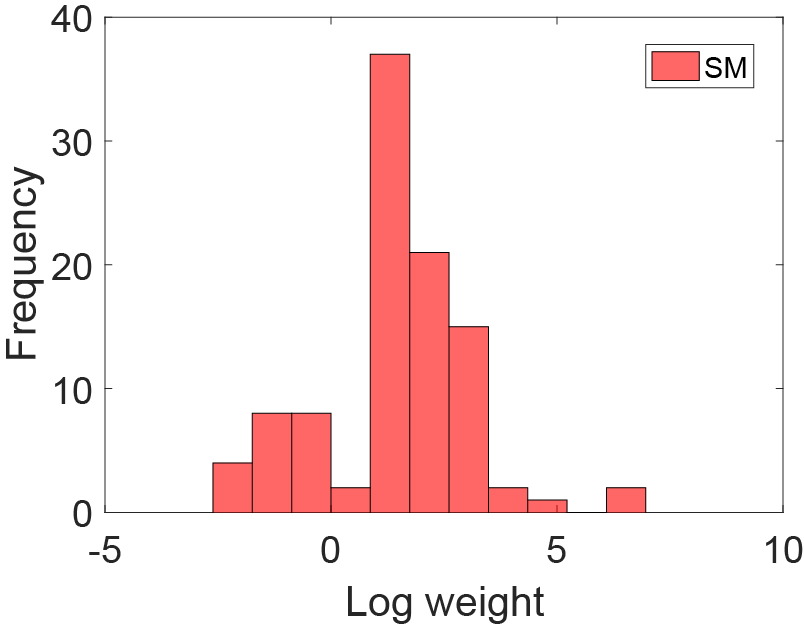} &
        \incpic{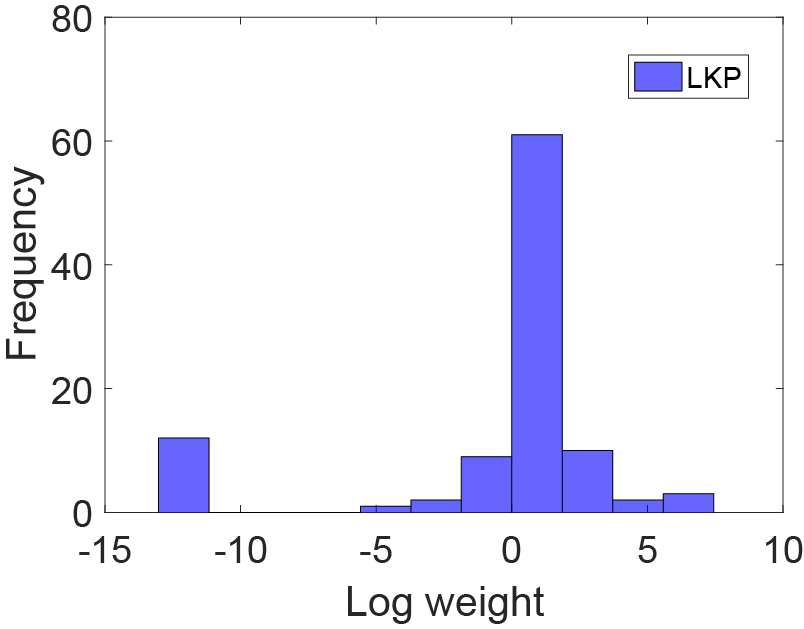} &
        \incpic{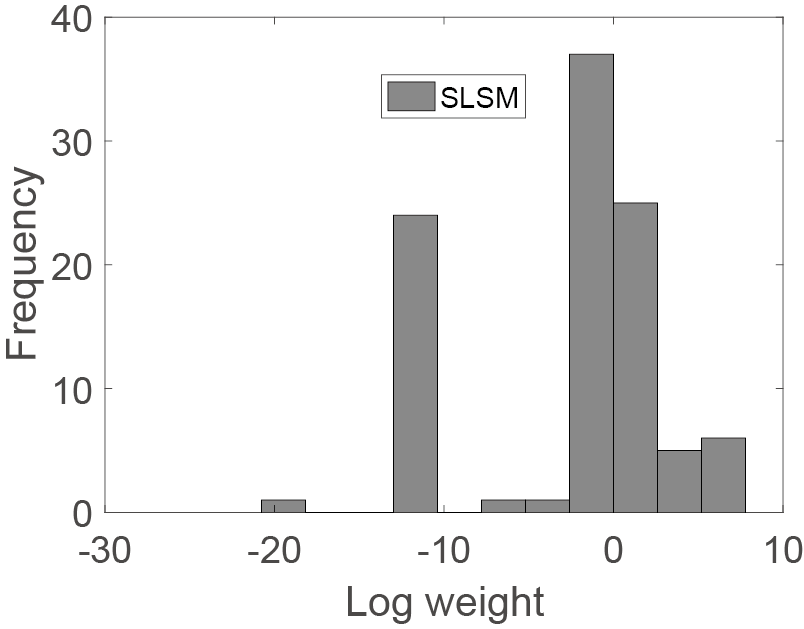}
        \\
        \incpic{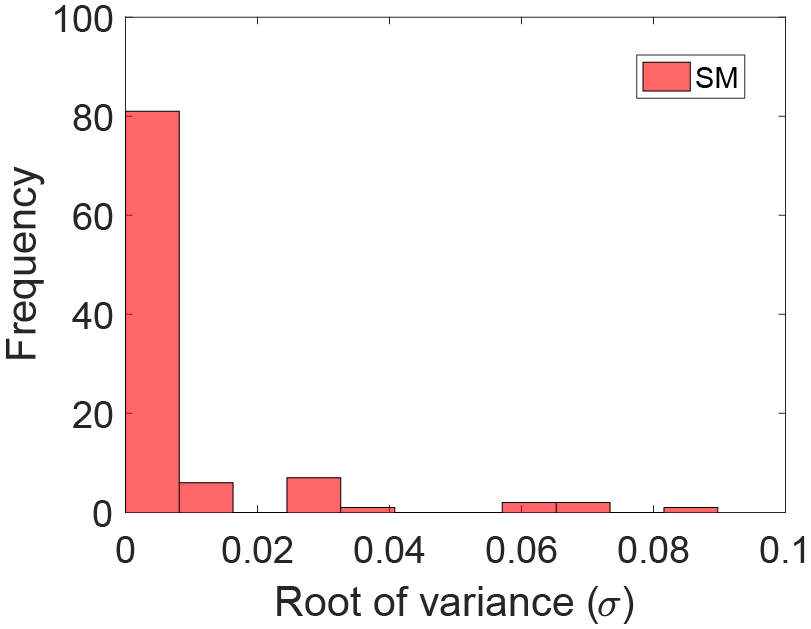} &
        \incpic{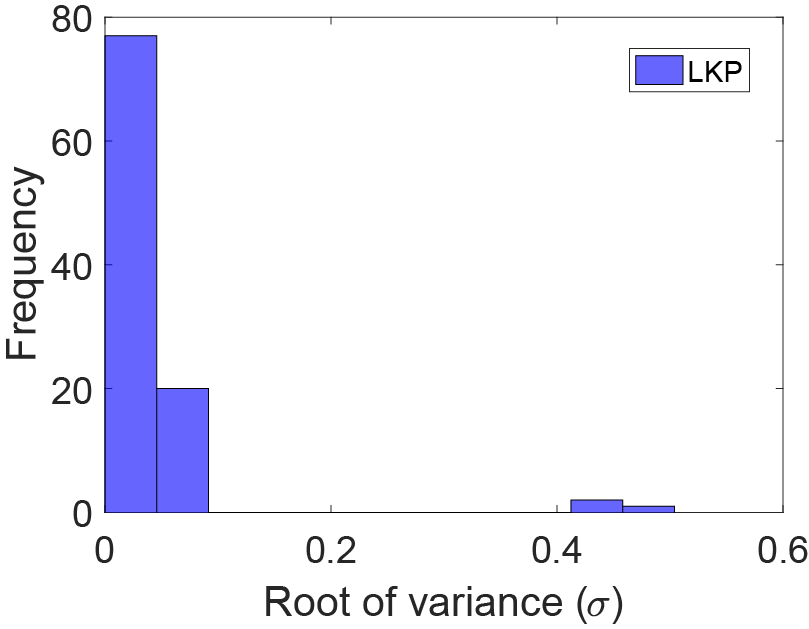} &
        \incpic{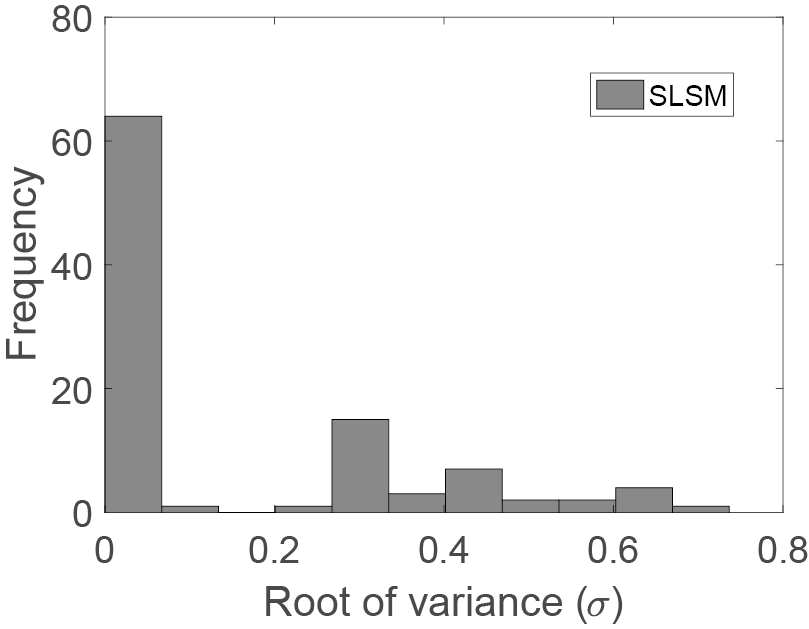}
        \\
        {(a) SM} &
        {(b) LKP} &
        {(c) SLSM}
    \end{tabular}
    \caption{Extrapolation and histogram of hyper-parameters of SM, LKP, and SLSM with 100 components performed on mean sunspot dataset.
    First row: extrapolation of SM, LKP, SLSM.
    Second row: histogram of log weight $\log(w)$;
    Third row:  histogram of variance $\sigma$;
    For histogram of weights in LKP and SLSM, most of the components have very small weights close to zero, which means the LKP and SLSM are very sparse. While for SM a large number of components have weights bigger than 10, which means SM is very dense and these SM components usually have bigger magnitude in the frequency domain.
    On the other hand, the histogram of variance $\sigma$ shows that SLSM components are much more heavy tailed with bigger variances than those of LKP and SM.
    SM components in the frequency domain are very short tailed, which exhibit fast decaying behaviors.
    }
    \label{fig:sun-spot-hist}
\end{figure*}

%----------------------------------------------------------------------------------------
\subsection{Long-term prediction of mean sunspot numbers: large number of kernel components}\label{sec:exp_sun}
%----------------------------------------------------------------------------------------

Here we further assess long-term forecasting capability of the considered kernels, this time with a large number of components. We consider the mean total sunspot number dataset.
For this dataset the monthly number of sunspots was collected from Jan. 1842 - Dec. 1933 at the Royal Observatory of Belgium, Brussels, resulting in 1104 time points. We consider the 13-month smoothed data \footnote{\url{http://www.sidc.be/silso/datafiles}}.
We consider $Q=100$  for the SM, LKP and SLSM kernels.
In order to increase the difficulty of extrapolation and perform longer-term forecasting, we use less training samples (45$\%$) than test samples (55$\%$). 

Results are shown in Figure~\ref{fig:sun-spot-hist}, in particular 
  the hyper-parameters histogram of SM, LKP, and SLSM kernels, which  indicates that in addition to long-range covariance, skewness is also beneficial for  robust learning in the presence of a large number of components. 
 However a large value of $Q$  affects the optimization time due to the larger of hyper-parameter space. 
Table~\ref{tab:MSE} and Table~\ref{tab:MAE} shows improved MSE and MAE results of SLSM over the other kernels, while in this experiment NLML of SLSM is the smallest (see Table~\ref{tab:NLML}). 

%----------------------------------------------------------------------------------------
\subsection{Extrapolating CPU relative performance: multidimensional computer hardware data}\label{sec:exp_comp}
%----------------------------------------------------------------------------------------

We now investigate the performance of the considered spectral kernels on an extrapolation task with multi-variate data.
We consider a multidimensional computer hardware dataset. 
The dataset recorded relative CPU performance collected on Oct. 1987. There are 209 multivariate samples and the following 8 features: vendor name, model name, machine cycle time, minimum main memory, maximum main memory, cache memory, minimum channels, maximum channels. 
The task is to predict CPU relative performance. In this experiment we use $Q=10$ components for SLSM, SLSM($\gamma=0$), and SM. 
We use the first $60\%$ of the data for training and the rest for testing. Before training the GPs, we normalize the training data so that the data is centered around the mean and has unit standard deviation. The mean and standard deviation of the training data are used to normalized test data.

Note that for SM, LKP, and SLSM on this unevenly discrete sampled multi-dimensional dataset, the spectral density structures are unclear and their hyper-parameter spaces are much larger than for time series. Input locations in this multi-dimensional dataset are not like those of time series: they are unevenly discretely  distributed. Thus we cannot get the multidimensional Fourier transform of the data. So we do not know their spectral densities structure. The size of their hyper-parameter space is equal to their input dimension times the length of the time series. 
Therefore, on this small multi-dimensional dataset we did not apply our procedure for pruning components.  Also, we could not use informed hyper-parameter initialization for SM, LKP, and SLSM, hence we just randomly initialize all kernels.

LKP with L\'evy prior and RJ-MCMC inference was developed and tested on one-dimensional data. So, in order to compare LKP with SLSM we only consider SLSM with $\gamma=0$, with standard exact inference instead of RJ-MCMC inference used in LKP.
The other baseline kernels use automatic relevance determination (ARD) to scale each dimension independently. 
Results of this experiment, in terms of MSE, MAE and NLML are given in Table~\ref{tab:MSE}, Table~\ref{tab:MAE}, and Table~\ref{tab:NLML}, respectively.

%----------------------------------------------------------------------------------------
\subsection{Long-term telecommunication  forecasting: a large dataset}\label{large-dataset}
%----------------------------------------------------------------------------------------

In order to investigate scalability of  SLSM, we consider a large dataset: the pole telecom dataset \cite{nguyen2014fast}. The pole telecomm dataset describing a telecommunication problem has 26 non-constant input dimensions, 10000/5000 samples for train/test, respectively.
By using rBCM, we can reduce the computational complexity of GP inference to $\compO(mn_{m}^{2})$, where $m$ and $n_{m}$ denote the number of BCM's and the size of data used in each BCM, respectively.

For SLSM, SLSM($\gamma=0$), and SM, we use $Q=5$ components and randomly hyper-parameter initialization due to the unclear spectral density structure. 
ARIMA, ETS, and LKP using L\'evy prior and RJ-MCMC inference cannot be (directly) applied to multi-dimensional data.
In this experiment we do not perform automatic pruning of components with the LTH algorithm, due to the relatively low efficiency of its inference process.
Results indicate good performance of SLSM when applied to a large dataset.

\subsection{Long-term forecasting: large number of times series}

We conclude our empirical investigation with experiments on the M3-Competition monthly data, which includes more than 1000 time series \cite{makridakis2000the}. 
The M3-Competitions data consists mainly of business and economic time series of five types: micro, industry, macro, finance, and demographic.
Because most of these time series are short, in order to perform long-term forecasting,  we use the first $80\%$ time points of each time series as training data, and the remaining data of each time series as test.   

Time series in M3-Competition monthly data have different overall scales. The MSE is sensitive to the overall scale of the test data.  Therefore, here we use the Standardized Mean Squared Error (SMSE): $\mathrm {SMSE}(\vy_{i}^{*}, \tilde{\vy}_{i}^{*}) = {\sum _{i=1}^{n}\big(y_{i}^{*}-\tilde{y}_{i}^{*}\big)^{2}}/({n\cdot\mathbb{V}[{{\vy}^{*}}]})$ by normalizing with the variance of the target values of the test \cite{Rasmussen2006}, then average the predictive performances of all M3-Competition monthly time series. 

Table~\ref{tab:SMSE-bench} contains SMSE results obtained by GP with spectral kernels,  and by the ARIMA, and ETS methods. Results indicate superior performance of GP with spectral kernels over traditional forecasting methods. Note, however, that we did not tune the parameters of ARIMA, and ETS methods, but used their default values given in ECOTOOL toolbox.

\newcommand{\std}[1]{{\scalebox{0.8}{\hspace{1mm}$\pm$\hspace{1mm}#1}}}
\begin{table*}[ht]
	\caption{Performance of SLSM and other kernels in terms of average MSE (over 10 independent runs) $\pm$ standard deviation, and performance of traditional forecasting methods.}\label{tab:MSE}
	\centering
    \scriptsize
    \tabcolsep=0.08cm
    \scalebox{1}{
	\begin{tabular}{l r @{} l r @{} l r @{} l r @{} l r @{} l }
		\toprule
		{Kernel} &  \multicolumn{2}{c}{Rail miles} & \multicolumn{2}{c}{Electricity} & 		\multicolumn{2}{c}{Sunspot} & \multicolumn{2}{c}{Computer} & \multicolumn{2}{c}{Pole-telecom}\\
		\midrule
		LIN   & 6910&$\std{871}$ & 9710&$\std{616}$  & 4413&$\std{1296}$ & 1850&$\std{49}$ & 2397&$\std{24}$ \\
		SE   & 206248&$\std{98312}$ & 251272&$\std{3600}$  & 2125&$\std{71}$ & 36156&$\std{4740}$ & 58&$\std{14}$ \\
		Poly  & 304892&$\std{4891}$ & 260211&$\std{1832}$  & 2179&$\std{140}$ & 41472&$\std{2530}$ & 108&$\std{28}$ \\
		PER  & 9107&$\std{581}$ & 10788&$\std{5386}$  &2266&$\std{1153}$ & 25266&$\std{816}$ & 1614&$\std{1120}$ \\
		RQ  & 8433&$\std{2542}$ & 11495&$\std{1688}$  & 2175&$\std{115}$ & 5402&$\std{1174}$ & 51&$\std{7}$ \\
		MA & 259732&$\std{20192}$ &238185&$\std{7417}$  &2258&$\std{221}$ & 34239&$\std{5513}$ & 49&$\std{7}$  \\
		Gabor & 304877&$\std{4884}$ & 260535&$\std{2019}$  & 2236&$\std{148}$ & 42642&$\std{224}$ & 2612&$\std{0.9}$ \\
		NN   & 5159&$\std{333}$ & 9652&$\std{1783}$ & 1234&$\std{108}$ & 12239&$\std{1147}$ & 230&$\std{4}$ \\
		ADD  & 282873&$\std{15384}$ & 250704&$\std{3325}$  & 2265&$\std{178}$ & 676&$\std{85}$ & 57&$\std{7}$ \\
		ARIMA  & 7622& & 163950&  & 25031& & --& & --& \\ 
		Auto-ARIMA  & 290116& & 216497&  & 1662& & --& & --& \\
		ETS  & 3753& & 22721& & 514272& & --& & --& \\              
		SM  & 4527&$\std{571}$ &4049&$\std{400}$  & 795&$\std{31}$ & {559}&{}$\std{52}$ & 42&$\std{6}$ \\
		LKP & 5023&$\std{581}$ & 4257&$\std{424}$ & 613&$\std{74}$ & --& & --& \\
		SLSM$_{\gamma=0}$ & 4225&$\std{569}$ & 3476&$\std{197}$  & 385&$\std{32}$ & {538}&{}$\std{42}$ & 40&$\std{1}$ \\
		SLSM & 1466&$\std{220}$ & 3167&$\std{118}$ & 314&$\std{22}$ & \tb{298}&$\std{26}$ & \tb{29}&$\std{1}$ \\
		SM (LTH) & 3214&$\std{583}$ & 3686&$\std{214}$ & 772&$\std{31}$ & --& & --& \\
		SLSM$_{\gamma=0}$ (LTH) & 2552&$\std{698}$ & 3253&$\std{196}$ & 321&$\std{19}$ & --& & --& \\
		SLSM (LTH) & \tb{1205}&$\std{146}$ & \tb{3047}&$\std{104}$ & \tb{228}&$\std{37}$ & --& & --& \\
		\bottomrule
	\end{tabular}}
\end{table*}

\begin{table*}[ht]
    \scriptsize
	\caption{Performance of SLSM and other kernels in terms of average MAE (over 10 independent runs) $\pm$ standard deviation, and performance of traditional forecasting methods.}\label{tab:MAE}
	\centering
    \scalebox{1}{   
	\begin{tabular}{l r @{} l r @{} l r @{} l r @{} l r @{} l}
		\toprule
		{Kernel} &  \multicolumn{2}{c}{Rail miles} & \multicolumn{2}{c}{Electricity} & 		\multicolumn{2}{c}{Sunspot} &
		\multicolumn{2}{c}{Computer} & \multicolumn{2}{c}{Pole-telecom} \\
		\midrule
		LIN   & 65.8&\std{20.2} & 83.0&\std{2.9}  & 37.5&\std{0.9} & 33.1&\std{0.1} & 41.0&\std{0.2} \\
		SE   & 530.6&\std{22.1} & 467.7&\std{60.3}  & 27.4&\std{12.2} & 32.6&\std{13.6} & 3.8&\std{0.1} \\
		Poly  & 548.1&\std{4.8} & 502.1&\std{1.5}  & 39.2&\std{1.1} & 119.0&\std{3.2} & 4.2&\std{0.4} \\
		PER  & 84.6&\std{0.9} & 72.4&\std{26.5}  &35.5&\std{2.0} & 82.9&\std{7.5} & 25.9&\std{13.1} \\
		RQ  & 77.8&\std{11.5} & 81.3&\std{3.2}  & 36.9&\std{2.6} & 30.1&\std{12.3} & 3.7&\std{0.3} \\
		MA & 495.9&\std{29.9} &472.2&\std{10.8}  &37.4&\std{1.1} & 20.1&\std{4.1} & 3.4&\std{0.2} \\
		Gabor & 548.1&\std{4.8} & 501.1&\std{4.1}  & 37.8&\std{1.0} & 120.6&\std{1.8} & 29.3&\std{0.9} \\
		NN   & 67.9&\std{15.8} & 75.9&\std{3.1} & 29.0&\std{1.1} & 35.5&\std{0.6} & 11.6&\std{0.1} \\
		ADD  & 528.5&\std{17.5} & 489.7&\std{8.3}  & 38.0&\std{1.4} & 13.2&\std{2.3} & 4.2&\std{0.3} \\
		ARIMA  & 70.5& & 394.7&  & 103.4& & --& & --& \\
		Auto-ARIMA  & 534.5& & 456.6&  & 31.7& & --& & --& \\  
		ETS  & 49.9& & 111.9&  & 581.2& & --& & --& \\                  
		SM  & 56.2&\std{11.8} &50.5&\std{4.2}  & 24.3&\std{4.1} & 17.2&\std{4.8} & 3.6&\std{0.2} \\
		LKP & 64.3&\std{12.8} & 71.7&\std{9.2}  & 24.6&\std{6.1} & --& & --& \\
		SLSM$_{\gamma=0}$ & 54.8&\std{5.2} & 64.5&\std{12.9}  & 15.7&\std{3.1} & 12.0&\std{2.8} & 3.4&\std{0.1} \\
		SLSM & 41.8&\std{6.6} & 41.7&\std{3.5} & 12.4&\std{1.8} & \tb{9.4}&\std{1.5} & \tb{2.9}&\std{0.1} \\
		SM  (LTH) & 47.0&\std{4.4} & 44.1&\std{1.9} & 23.9&\std{0.9} & --& & --& \\
		SLSM$_{\gamma=0}$ (LTH) & 40.2&\std{5.6} & 41.6&\std{1.6} & 13.9&\std{0.5} & --& & --& \\
		SLSM  (LTH) & \tb{28.4}&\std{3.2} & \tb{40.8}&\std{0.9} & \tb{12.2}&\std{1.3} & --& & --& \\
		\bottomrule
	\end{tabular}}
\end{table*}

\begin{table*}[ht]
	\caption{Performance of SLSM and other kernels in terms of average NLML (over 10 independent runs) $\pm$ standard deviation, and performance of traditional forecasting methods.}\label{tab:NLML}
	\centering
    \tabcolsep=1.2mm
    \scriptsize
    \scalebox{1}{   
	\begin{tabular}{l r @{} l r @{} l r @{} l r @{} l r @{} l }
		\toprule
		{Kernel} & \multicolumn{2}{c}{Rail miles} & \multicolumn{2}{c}{Electricity} &
		\multicolumn{2}{c}{Sunspot} &
		\multicolumn{2}{c}{Computer} & \multicolumn{2}{c}{Pole-telecom} \\
		\midrule
		LIN  & 694.3&\std{101.0}  & 375.1&\std{50.1}  & 1336.7&\std{380.1} & 625.8&\std{14.5} & 49926&\std{22.4} \\
		SE   & 671.6&\std{151.1} & 339.3&\std{49.5} & 1784.0&\std{540.2} & 548.4&\std{94.3} & 40600&\std{1013.2} \\
		Poly  & 935.2&\std{135.2}  & 379.4&\std{67.0} & 2620.2&\std{546.6} & 622.5&\std{114.7} & 42670&\std{1351.9} \\
		PER  & 715.6&\std{64.4} & 359.6&\std{96.2}  &1702.4&\std{279.1}& 745.5&\std{3.4} & 48543&\std{3287.2} \\
		RQ  & 663.9&\std{96.4} &338.6&\std{46.0} & 1639.6&\std{576.9}& 257.6&\std{50.9} & 39058&\std{175.4} \\
		MA & 770.0&\std{78.7} &339.5&\std{77.4} &1898.7&\std{715.6}& 583.5&\std{61.7} & 39359&\std{509.4} \\
		Gabor & 936.7&\std{136.1} & 415.3&\std{79.0} & 2258.3&\std{603.3}& 824.8&\std{24.0} & 53433&\std{0.1} \\
		NN  & 744.3&\std{13.2} & 358.9&\std{58.0} & 1043.5&\std{171.8}& 420.2&\std{17.4} & 45119&\std{23.5} \\
		ADD & 766.0&\std{22.1} & 335.6&\std{68.5} & 791.9&\std{223.8}& 303.0&\std{4.5} & 39616&\std{182.6} \\
		SM & \tb{595.7}&\std{11.8} &337.7&\std{32.1} & 623.9&\std{11.2}& \tb{180.2}&\std{11.9} & 49679&\std{16.3} \\
		LKP & 699.3&\std{21.3} & 334.3&\std{5.0} & 612.1&\std{4.6}& --& & --& \\
		SLSM$_{\gamma=0}$ & 686.1&\std{23.0} & 339.5&\std{3.5} & 612.0&\std{3.0}& 217.1&\std{5.2} & 49630&\std{135.6} \\
        SLSM & 658.6&\std{14.5} &342.5&\std{3.6} & \tb{611.4}&\std{1.9}& 205.3&\std{5.2} & \tb{36812}&\std{997.2} \\
		SM (LTH) & 613.3&\std{45.7} & 337.9&\std{4.4} & 616.3&\std{6.9} & --& & --& \\
		SLSM$_{\gamma=0}$ (LTH) & 641.5&\std{44.2} & \tb{333.1}&\std{4.3} & 617.5&\std{1.1} & --& & --& \\
		SLSM (LTH) & 638.6&\std{23.4} & 340.0&\std{4.1} & 613.0&\std{2.2} & --& & --& \\	
		\bottomrule
	\end{tabular}}
\end{table*}

\begin{table*}[ht]
    \scriptsize
    \caption{Effect of the LTH pruning algorithm on GP models, in terms of percentage of kernel components pruned and increase in MAE and MSE performance.}\label{tab:prun-component}
    \centering
    \tabcolsep=1.6mm
    \scalebox{0.95}{    
    \begin{tabular}{l l r @{.} l r @{.} l r @{.} l r @{.} l r @{.} l}
        \toprule
        Dataset & Kernel & \multicolumn{2}{c}{Pruned nr.} & \multicolumn{2}{c}{Reduced hyps} & \multicolumn{2}{c}{Improved MSE} & \multicolumn{2}{c}{Improved MAE} \\
        \midrule
        Rail miles   & SM (LTH)           & {2}&{4} & {7}&{2}  & {29}&{01}$\%$ & {16}&{27}$\%$ \\
        Rail miles   & SLSM$_{\gamma=0}$ (LTH) & {3}&{5} & {10}&{5} & {49}&{19}$\%$ & {37}&{55}$\%$ \\
        Rail miles   & SLSM (LTH)         & {4}&{1} & {16}&{4} & {17}&{77}$\%$ & {32}&{15}$\%$ \\
        Sunspot      & SM (LTH)           & {4}&{7} & {14}&{1} & {2}&{84}$\%$  & {1}&{56}$\%$ \\
        Sunspot      & SLSM$_{\gamma=0}$ (LTH) & {5}&{1} & {15}&{3} & {47}&{71}$\%$ & {43}&{52}$\%$ \\
        Sunspot      & SLSM (LTH)         & {5}&{3} & {21}&{2} & {27}&{41}$\%$ & {2}&{24}$\%$ \\
        Electricity  & SM (LTH)           & {2}&{6} & {7}&{8}  & {8}&{95}$\%$  & {12}&{65}$\%$ \\
        Electricity  & SLSM$_{\gamma=0}$ (LTH) & {3}&{0} & {9}&{0}  & {23}&{58}$\%$ & {41}&{94}$\%$ \\
        Electricity  & SLSM (LTH)         & {2}&{8} & {11}&{2} & {3}&{78}$\%$  & {2}&{23}$\%$ \\            
        \bottomrule
    \end{tabular}}
\end{table*}

\begin{table}[!htb] 
    \caption{Performance in terms of SMSE on the M3-Competition monthly data.}\label{tab:SMSE-bench}
    \centering
    \scriptsize
    \tabcolsep=0.2cm
    \scalebox{1.0}{ 
        \begin{tabular}{l r @{.} l r @{.} l r @{.} l r @{.} l r @{.} l r @{.} l }
            \toprule
            {Kernel} &    \multicolumn{2}{c}{Micro} & \multicolumn{2}{c}{Industry} & \multicolumn{2}{c}{Macro} & \multicolumn{2}{c}{Finance} & \multicolumn{2}{c}{Demographic} \\
            \midrule            
            
            ARIMA  &  {4}&{71}  & {2}&{86} & {10}&{86} & {8}&{53}  & {13}&{60} \\  
            
            Auto-ARIMA  &  {41}&{43}  & {13}&{58} & {130}&{06} & {33}&{86}  & {72}&{05} \\
            
            ETS  & {3}&{05}  & {4}&{72} & {14}&{87} & {6}&{60}  & {18}&{16} \\
            
            SM &  {1}&{92} & {2}&{29} & {7}&{03} & {4}&{58} & {13}&{46}\\
            %        & {42}&{19}$\pm{5.67}$ \\
            
            LKP &  {3}&{71} & {4}&{34} & {7}&{75} & {6}&{57}  & {13}&{98} \\	
            
            SLSM$_{\gamma=0}$  & {1}&{76}  & {2}&{14} & {4}&{41} & {3}&{46}  & {9}&{23} \\
            
            SLSM & {1}&{70} & {1}&{97} & {3}&{81} & {3}&{04}  & {8}&{47}\\		
            
            SM (LTH)  &{1}&{71}  & {2}&{06} & {6}&{98} & {4}&{26}  & {13}&{31} \\

            SLSM$_{\gamma=0}$  (LTH) &  {1}&{39}  & {1}&{87} & {4}&{37} & {3}&{39}  & {8}&{33} \\

            SLSM  (LTH)&  \tb{1}&\tb{23} & \tb{1}&\tb{63} & \tb{3}&\tb{76} & \tb{2}&\tb{93}  & \tb{7}&\tb{76} \\            
            \bottomrule
    \end{tabular}}
\end{table}

%----------------------------------------------------------------------------------------
\subsection{Discussion}
%----------------------------------------------------------------------------------------

Results of our experiments indicate that a GP with the SLSM kernel consistently outperforms other baselines in terms of MSE and MAE, but its NLML is not always the lowest.
A reason for this phenomenon could be that the marginal likelihood surfaces for the GPs with SM, LKP, and SLSM kernels usually have many local optima. 
Also, computation of NLML (Eq~\eqref{eq:nlml}) uses the training data, which may show a good fit, but this does not guarantee a better generalization performance.
Analogously, our experiments indicate that 
a better fitting of the empirical spectral density does not necessarily yield the best extrapolation performance (see Figures~\ref{fig:elec} and \ref{fig:rail}). 
Based on these considerations, MSE and MAE are better evaluation metrics, because they are computed on the test data, with a slight preference for MAE, which is less sensitive to outliers.

Unsatisfactory performance of methods like SE, Poly, MA, Gabor, ADD, and Auto-ARIMA, is due to the fact that these methods are designed to perform interpolation, while we consider extrapolation tasks. Better performance of GPs with SLSM over other spectral kernels like SM and LKP, can be explained by the fact that SLSM  extends the covariance range in the GP model, which can correlate test points far away from training and thus be beneficial for long range extrapolation. Specifically, compared to other kernels, the spectral density of the SLSM kernel tends to have very sharp peaks.
In addition, we can observe that the spectral density SLSM has much bigger variance than SM and LKP, which further extends the tails of components.

Since the variance in the frequency domain is inversely related to the length-scale in the time domain, these components will lead to a kernel with long-range covariances in the time domain. This can explain the good extrapolation performance of the kernel. Also, compared to the SM and LKP kernels, the SLSM kernel can discover more short period patterns, corresponding to spectral density components with large means. In general the long-term variation is more important for extrapolation because short period patterns mixed with noise and specific local factors are difficult to capture.

Note that  SLSM ($\gamma=0$) has the same kernel structure and implementation as LKP, but uses LBFGS implemented in GPML toolbox to perform inference instead of Reversed-Jump MCMC (RJ-MCMC) in LKP \cite{Jang2017}. 
Results of our experiments indicate that SLSM is also superior to SLSM($\gamma=0$), showing that the kernel used in the LKP method (that is, SLSM($\gamma=0$)) can be improved by incorporating skewness.  
Also, results show superior performance of SLSM($\gamma=0$) over LKP, indicating that 
for this type of kernel, gradient descent optimization usually infers a GP model that is better than that obtained using sampling based optimization.
In general, we could not compare results achieved independently by the authors of the LKP method, because no quantitative results are reported in their paper. For SM, In general, we could not compare
results achieved independently by the authors of the LKP method \cite{Jang2017} , because no quantitative results are reported in their paper. For SM, we run experiments also on the
monthly airline passenger with same training and test splitting as in \cite{Wilson2013}.  The monthly airline passenger time series has length 144, with measurements from 1949 to 1961. It uses  96 records for training and the rest 48 records as test data.
For SM we achieved the following results: average NLML, MAE, and  MSE equal to 353.99,	16.87, and	460.04, respectively, close to those achieved independently in \cite{Wilson2013}. Thus our implementation of SM is in agreement with the original method described in \cite{Wilson2013}.  On the same dataset, SMSL achieved the following results: average NLML, MAE, and  MSE equal to 354.71, 16.78,	456.24, respectively.

We  performed experiments with varying number of components, $Q=10$, $Q$ between $2$ and $10$, $Q=100$, and investigated the effect of pruning components using the proposed LTH algorithm. Results indicate that
automatic pruning using our LTH algorithm can drastically reduce the number of kernel components, which reduces the size of the hyper-parameter space, and hence improves performance (see Table \ref{tab:prun-component}).

% ============================================================================================
\section{Conclusion}\label{sec:conclude}
% ============================================================================================
We have proposed a new kernel for long-term forecasting with GPs, derived in a principled way by modeling the empirical spectral densities using a skewed Laplace spectral mixture, whose characteristics are more beneficial for long-range forecasting  than those of the Gaussian mixture used in SM kernels. 

Experiments on real-world datasets demonstrated that by using the SLSM kernel dense structures as well as sparse structures in the data can be captured. In addition, the SLSM kernel was shown to be able to learn periodical patterns, especially short-term patterns in the signal. 

Overall, the SLSM kernel appears to be especially beneficial for time series containing patterns with long-range dependencies, as indicated, for example, by experiments with the electricity and rail miles datasets. 
Other real-world tasks \cite{5178729} suited for SLSM include forecasting sea level  \cite{ercan2013long} to estimate the climate change trends, and long-term forecasting of remaining useful battery life \cite{richardson2017gaussian} to ensure reliable system operation and to minimise maintenance costs.

The SLSM kernel also has a connection with RQ kernel, which not only provides a new interpretation of the RQ kernel, but also demonstrates the usefulness of the SLSM kernel as a principled alternative to the exponential and RQ kernels.
We might further extend SLSM to increase its long-range modeling capability by using a RQ kernel with $\alpha<{1}$ instead of the Cauchy function term of the SLSM kernel.

The SLSM kernel can also be  generalized to scalable inference like 
Cartesian structured multidimensional datasets like images following \cite{Gilboa2015,Wilson2014,Wilson2014a,Wilson2015}. The multidimensional SLSM kernel becomes:
\begin{align}\label{eq:lsm-md}
{k}_{\lsm}({\tau})
&=\prod_{p=1}^{P}\sum_{i=1}^{Q}{k}_{\lsm,i}^{(p)}({\tau})
\end{align}
where ${k}_{\lsm,i}^{(p)}({\tau})$ is the $i$-th SLSM component on the $p$-th dimension. 
Equation \eqref{eq:lsm-md} can be used for fast inference by decomposing ${K}_{\lsm}$ as the Kronecker product of matrices over each input dimension ${K}_{\lsm}={K}_{\lsm}^{1}\otimes{...}\otimes{K}_{\lsm}^{P}$.
It is interesting to investigate applications involving such data.

The main limitation of the proposed kernel is that it is not suited for modeling  non-stationary phenomena \cite{Remes2017}. An interesting open problem is exploring how to overcome this limitation without loss of interpretation. 

\section*{Acknowledgements}
The work was supported by the Radboud University and the China Postdoctoral Science Foundation (No. 2020M671899).
%\section*{acknowledgements}

% \section{References}
% \bibliography{config/LaplaceSM}
% \bibliographystyle{config/elsarticle-num}
% \bibstyle:{styles/elsarticle-num}
% \bibliographystyle{styles/aaai}

%\bibliographystyle{config/spbasic}      % basic style, author-year citations
%\bibliographystyle{config/spphys}       % APS-like style for physics
%\bibliography{Asymmetrical_Laplace}   % name your BibTeX data base
%\bibliographystyle{unsrt}
%\bibliography{LaplaceSM}   % name your BibTeX data base
\bibliographystyle{unsrt}

\end{document}